\PassOptionsToPackage{table}{xcolor}
\documentclass[10pt,twocolumn,letterpaper]{article}

\usepackage[pagenumbers]{iccv} 
\usepackage[accsupp]{axessibility}  

\usepackage{graphicx}
\usepackage{booktabs}
\usepackage{amsmath}
\usepackage{amssymb}
\usepackage{bm}
\usepackage{booktabs}
\usepackage{multirow}
\usepackage{makecell}
\usepackage{floatrow}
\usepackage{nicematrix}
\usepackage{array}
\newfloatcommand{capbtabbox}{table}[][\FBwidth]
\definecolor{iccvblue}{rgb}{0.21,0.49,0.74}
\usepackage[pagebackref,breaklinks,colorlinks,allcolors=iccvblue]{hyperref}

\def\paperID{****} 
\def\confName{ICCV}
\def\confYear{2025}

\title{Multi-view Gaze Target Estimation}

\author{
Qiaomu Miao\textsuperscript{1},
Vivek Raju Golani\textsuperscript{1},
Jingyi Xu\textsuperscript{1},
Progga Paromita Dutta\textsuperscript{1},
Minh Hoai\textsuperscript{2},
Dimitris Samaras\textsuperscript{1} \\
\textsuperscript{1}Stony Brook University \qquad
\textsuperscript{2}	The University of Adelaide \qquad \\
}
\begin{document}
\def\mA{\mathcal{A}}
\def\mB{\mathcal{B}}
\def\mC{\mathcal{C}}
\def\mD{\mathcal{D}}
\def\mE{\mathcal{E}}
\def\mF{\mathcal{F}}
\def\mG{\mathcal{G}}
\def\mH{\mathcal{H}}
\def\mI{\mathcal{I}}
\def\mJ{\mathcal{J}}
\def\mK{\mathcal{K}}
\def\mL{\mathcal{L}}
\def\mM{\mathcal{M}}
\def\mN{\mathcal{N}}
\def\mO{\mathcal{O}}
\def\mP{\mathcal{P}}
\def\mQ{\mathcal{Q}}
\def\mR{\mathcal{R}}
\def\mS{\mathcal{S}}
\def\mT{\mathcal{T}}
\def\mU{\mathcal{U}}
\def\mV{\mathcal{V}}
\def\mW{\mathcal{W}}
\def\mX{\mathcal{X}}
\def\mY{\mathcal{Y}}
\def\mZ{\mathcal{Z}} 

\def\bbN{\mathbb{N}} 
\def\bbR{\mathbb{R}} 
\def\bbP{\mathbb{P}} 
\def\bbQ{\mathbb{Q}} 
\def\bbE{\mathbb{E}}

\def\1n{\mathbf{1}_n}
\def\0{\mathbf{0}}
\def\1{\mathbf{1}}

\def\A{{\bf A}}
\def\B{{\bf B}}
\def\C{{\bf C}}
\def\D{{\bf D}}
\def\E{{\bf E}}
\def\F{{\bf F}}
\def\G{{\bf G}}
\def\H{{\bf H}}
\def\I{{\bf I}}
\def\J{{\bf J}}
\def\K{{\bf K}}
\def\L{{\bf L}}
\def\M{{\bf M}}
\def\N{{\bf N}}
\def\O{{\bf O}}
\def\P{{\bf P}}
\def\Q{{\bf Q}}
\def\R{{\bf R}}
\def\S{{\bf S}}
\def\T{{\bf T}}
\def\U{{\bf U}}
\def\V{{\bf V}}
\def\W{{\bf W}}
\def\X{{\bf X}}
\def\Y{{\bf Y}}
\def\Z{{\bf Z}}

\def\a{{\bf a}}
\def\b{{\bf b}}
\def\c{{\bf c}}
\def\d{{\bf d}}
\def\e{{\bf e}}
\def\f{{\bf f}}
\def\g{{\bf g}}
\def\h{{\bf h}}
\def\i{{\bf i}}
\def\j{{\bf j}}
\def\k{{\bf k}}
\def\l{{\bf l}}
\def\m{{\bf m}}
\def\n{{\bf n}}
\def\o{{\bf o}}
\def\p{{\bf p}}
\def\q{{\bf q}}
\def\r{{\bf r}}
\def\s{{\bf s}}
\def\t{{\bf t}}
\def\u{{\bf u}}
\def\v{{\bf v}}
\def\w{{\bf w}}
\def\x{{\bf x}}
\def\y{{\bf y}}
\def\z{{\bf z}}

\def\balpha{\mbox{\boldmath{$\alpha$}}}
\def\bbeta{\mbox{\boldmath{$\beta$}}}
\def\bdelta{\mbox{\boldmath{$\delta$}}}
\def\bgamma{\mbox{\boldmath{$\gamma$}}}
\def\blambda{\mbox{\boldmath{$\lambda$}}}
\def\bsigma{\mbox{\boldmath{$\sigma$}}}
\def\btheta{\mbox{\boldmath{$\theta$}}}
\def\bomega{\mbox{\boldmath{$\omega$}}}
\def\bxi{\mbox{\boldmath{$\xi$}}}
\def\bnu{\mbox{\boldmath{$\nu$}}}                                  
\def\bphi{\mbox{\boldmath{$\phi$}}}
\def\bmu{\mbox{\boldmath{$\mu$}}}

\def\bDelta{\mbox{\boldmath{$\Delta$}}}
\def\bOmega{\mbox{\boldmath{$\Omega$}}}
\def\bPhi{\mbox{\boldmath{$\Phi$}}}
\def\bLambda{\mbox{\boldmath{$\Lambda$}}}
\def\bSigma{\mbox{\boldmath{$\Sigma$}}}
\def\bGamma{\mbox{\boldmath{$\Gamma$}}}
                                  
\newcommand{\myprob}[1]{\mathop{\mathbb{P}}_{#1}}

\newcommand{\myexp}[1]{\mathop{\mathbb{E}}_{#1}}

\newcommand{\mydelta}[1]{1_{#1}}

\newcommand{\myminimum}[1]{\mathop{\textrm{minimum}}_{#1}}
\newcommand{\mymaximum}[1]{\mathop{\textrm{maximum}}_{#1}}    
\newcommand{\mymin}[1]{\mathop{\textrm{minimize}}_{#1}}
\newcommand{\mymax}[1]{\mathop{\textrm{maximize}}_{#1}}
\newcommand{\mymins}[1]{\mathop{\textrm{min.}}_{#1}}
\newcommand{\mymaxs}[1]{\mathop{\textrm{max.}}_{#1}}  
\newcommand{\myargmin}[1]{\mathop{\textrm{argmin}}_{#1}} 
\newcommand{\myargmax}[1]{\mathop{\textrm{argmax}}_{#1}} 
\newcommand{\myst}{\textrm{s.t. }}

\newcommand{\denselist}{\itemsep -1pt}
\newcommand{\sparselist}{\itemsep 1pt}

\definecolor{pink}{rgb}{0.9,0.5,0.5}
\definecolor{purple}{rgb}{0.5, 0.4, 0.8}   
\definecolor{gray}{rgb}{0.3, 0.3, 0.3}
\definecolor{mygreen}{rgb}{0.2, 0.6, 0.2}

\newcommand{\cyan}[1]{\textcolor{cyan}{#1}}
\newcommand{\blue}[1]{\textcolor{blue}{#1}}
\newcommand{\magenta}[1]{\textcolor{magenta}{#1}}
\newcommand{\pink}[1]{\textcolor{pink}{#1}}
\newcommand{\green}[1]{\textcolor{green}{#1}} 
\newcommand{\gray}[1]{\textcolor{gray}{#1}}    
\newcommand{\mygreen}[1]{\textcolor{mygreen}{#1}}    
\newcommand{\purple}[1]{\textcolor{purple}{#1}}       

\definecolor{greena}{rgb}{0.4, 0.5, 0.1}
\newcommand{\greena}[1]{\textcolor{greena}{#1}}

\definecolor{bluea}{rgb}{0, 0.4, 0.6}
\newcommand{\bluea}[1]{\textcolor{bluea}{#1}}
\definecolor{reda}{rgb}{0.6, 0.2, 0.1}
\newcommand{\reda}[1]{\textcolor{reda}{#1}}

\def\changemargin#1#2{\list{}{\rightmargin#2\leftmargin#1}\item[]}
\let\endchangemargin=\endlist
                                               
\newcommand{\cm}[1]{}

\newcommand{\vect}[1]{\mathbf{#1}}
\newcommand{\mtodo}[1]{{\color{red}$\blacksquare$\textbf{[TODO: #1]}}}
\newcommand{\myheading}[1]{\vspace{0.5ex}\noindent \textbf{#1}}
\newcommand{\htimesw}[2]{\mbox{$#1$$\times$$#2$}}


%
%
%

\newcommand{\Sref}[1]{Sec.~\ref{#1}}
\newcommand{\Eref}[1]{Eq.~(\ref{#1})}
\newcommand{\Fref}[1]{Fig.~\ref{#1}}
\newcommand{\Tref}[1]{Table~\ref{#1}}

\maketitle
\begin{abstract}
This paper presents a method that utilizes multiple camera views for the gaze target estimation (GTE) task. The approach integrates information from different camera views to improve accuracy and expand applicability, addressing limitations in existing single-view methods that face challenges such as face occlusion, target ambiguity, and out-of-view targets. Our method processes a pair of camera views as input, incorporating a Head Information Aggregation (HIA) module for leveraging head information from both views for more accurate gaze estimation, an Uncertainty-based Gaze Selection (UGS) for identifying the most reliable gaze output, and an Epipolar-based Scene Attention (ESA) module for cross-view background information sharing. This approach significantly outperforms single-view baselines, especially when the second camera provides a clear view of the person's face. Additionally, our method can estimate the gaze target in the first view using the image of the person in the second view only, a capability not possessed by single-view GTE methods. Furthermore, the paper introduces a multi-view dataset for developing and evaluating multi-view GTE methods.  Data and code are available at \url{https://www3.cs.stonybrook.edu/~cvl/multiview_gte.html}.
\end{abstract}    
\section{Introduction}
\label{sec:intro}

Gaze Target Estimation (GTE) is an important problem with applications in areas such as social behavior analysis~\cite{emery2000eyes, sakita2004flexible}, human-machine interactions~\cite{jacob2003eye, admoni2017social}, and mental disorder diagnosis~\cite{volkmar1990gaze, frazier2017meta}. Earlier works studied gaze behaviors using specialized equipment like eye trackers \cite{fathi2012learning, liu20203d} or head-mounted cameras \cite{park20123d, fathi2012social}, which are expensive and intrusive.  
Recent advances in deep learning \cite{resnet, ho2020denoising, vaswani2017attention, alexnet} have facilitated the development of GTE models to estimate gaze in the wild using ordinary scene cameras, thereby broadening their range of applications.

\begin{figure}[t]
\centering
\includegraphics[width=\linewidth]{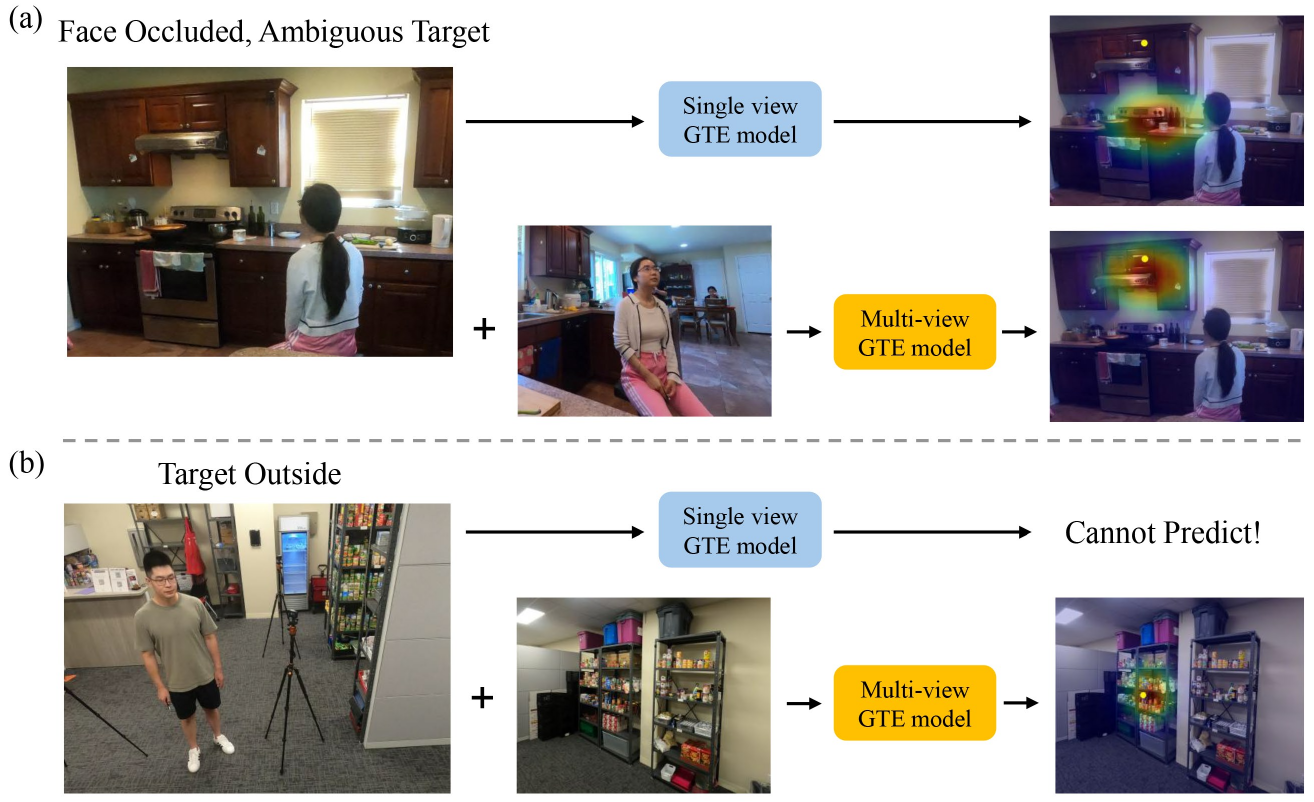}
\vspace{-0.2cm}
\caption{Benefits of multi-view GTE. Single-view GTE models struggle with input where the subject face is occluded, and cannot predict the target location outside the frame. In contrast, a multi-view GTE model can leverage another view's information to improve GTE accuracy, and predict the gaze target across views.} 
\label{fig:teaser} \vspace{-0.3cm}
\end{figure}

Several methods have been proposed for GTE using ordinary scene cameras. However, as shown in Fig.\ref{fig:teaser}, existing methods struggle with images in which the subject's face is not visible to the camera and multiple potential targets exist.  In addition, single view GTE methods only function when both the person and their gaze target are visible in the image. If the gaze target is outside the image, these methods cannot function at all. 
These limitations make current methods restrictive due to the limited field of view of a camera. 

Using multiple cameras provides a solution to these limitations. Compared to single-camera systems, multi-camera setups provide broader coverage and multiple perspectives of both human faces and the scene background. This enables more accurate gaze estimation by providing a clearer view of the face, and allows predicting gaze targets that may appear in a different camera view from the subject. Furthermore, multi-camera setups are widely utilized in many environments, such as supermarkets and lecture halls, that could benefit from non-intrusive gaze target estimation (GTE).


However, there are challenges in developing a method that effectively leverage multiple cameras for GTE. Due to perspective changes between different views, directly combining input images or extracted features without accounting for the geometric relationship between them, is not beneficial. Meanwhile, in real-world applications, explicit and complete 3D reconstruction of the scene and subject is not always feasible, as the views often have limited or no overlap. Even if 3D reconstruction is possible, performing it for every input---particularly when the subject or other people move in the scene---would be memory and time-intensive.

 In this paper, we introduce the first multi-view GTE model that effectively and efficiently leverages information from multiple camera views.
 Our model builds upon a transformer-based single-view GTE framework and extends it to process a pair of camera views. It incorporates a Head Information Aggregation (HIA) module that leverages the head appearance information and the geometric relationship between both views to enhance the head embedding and improve gaze direction estimation. The estimated gaze vectors are then processed by an Uncertainty-based Gaze Selection (UGS) module, which selects the more reliable gaze vector from the two views, replacing the predicted vector in the less reliable view. Additionally, the Epipolar-based Scene Attention (ESA) module integrates scene background information from different perspectives. Altogether, these modules enhance the GTE capability with both the gaze and scene information from multiple views.

Compared to single-view methods, our model leverages an additional view to achieve significant improvements, especially when the additional view captures the person's head appearance. Our model also addresses a unique scenario that existing single-view GTE models cannot handle: when the gaze target is visible in one view but the person is only visible in the second view. This scenario is challenging as triangulation cannot be used to infer the absolute depth of the person and scene due to little view overlap. 
To address this issue, we estimate absolute depth by comparing monocular depth maps against a pre-reconstructed 3D scene, generated from a multi-view reconstruction model \cite{wang2024dust3r} prior to training. This allows us to estimate the absolute depth for all new inputs in a scene by applying the 3D reconstruction model only once.

To train and evaluate our method, we introduce the first dataset for multi-view GTE: the Multi-View Gaze Target (MVGT) dataset. This dataset was collected across four real-world scenes, featuring 28 subjects with precise gaze target annotations. The images were captured simultaneously from multiple calibrated cameras positioned in the scene. We also introduce a data collection protocol that can collect data non-intrusively and obtain precise gaze targets without introducing artifacts.

In summary, our main contributions are: (1) the first exploration of the multi-view GTE task; (2) a novel GTE model that effectively leverages the human head and scene information from multiple views, surpassing current state-of-the-art single-view models; (3) the MVGT dataset, featuring images captured synchronously by multiple calibrated cameras with precise target annotations, along with a data collection protocol to collect accurate gaze target annotations without creating artifacts.

\section{Related Works}
\label{sec:formatting}

\myheading{Gaze target estimation} was first investigated in \cite{NIPS2015_ec895663}, which introduced the GazeFollow dataset. \citet{lian2018believe} generated 2D direction fields for GTE, while Chong et al. extended the task to predicting whether the target is located in the image \cite{chong2018connecting} and GTE in videos \cite{chong2020detecting}. Later models adopted additional modalities such as depth \cite{fang2021dual, Bao_2022_CVPR, tonini2022multimodal, miao2023patch, yang2024gaze} and human pose \cite{Bao_2022_CVPR, gupta2022modular, yang2024aaai} for improvement. Recent methods compute 3D field-of-view (FoV) heatmap as gaze target priors by predicting a 3D gaze vector and using depth input \cite{tafasca2023childplay, hu2022we, gf3d_cvpr23}. Additional works leveraged transformer architectures for GTE \cite{song2024vitgaze, tafasca2024sharingan}, and jointly predicting the head location and gaze targets \cite{tu2022end,tonini2023object}. Some works explored unified GTE and social gaze prediction \cite{gupta2024unified, gupta2024mtgs}, or GTE using fewer labels \cite{miao2024diffusion, tonini2024gtd}. All these methods focus on GTE in a single camera view. {Several works have explored solutions for out-of-frame gaze targets. \citet{recasens2017following} introduced a dataset and a model to predict gaze targets in future video frames based on a person seen in the current frame. \citet{gazefollowing360} investigated GTE in 360° images, and \citet{yu2019see} learned a joint embedding for first- and third-person frames. However, they did not leverage the explicit geometric relationships between views from the calibrated camera parameters. Furthermore, no previous work has explored GTE with multiple third-person views, which is increasingly prevalent in real-world applications.}

\begin{figure*}[ht]
    \centering
    \includegraphics[width=0.98\linewidth]{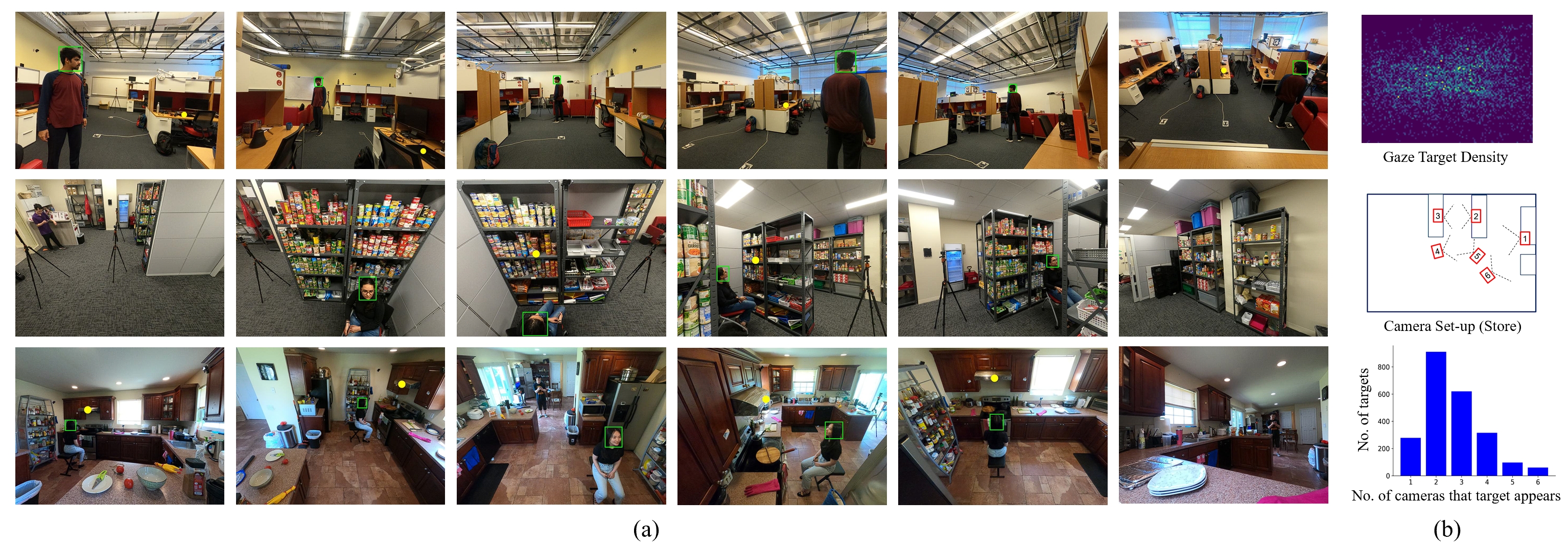}
    \vspace{-0.3cm}
    \caption{Dataset samples and information. (a) Example images and annotations of the subject's head location (green bounding box) and the gaze target (yellow dot) from all 6 cameras. (b) Dataset information including the density of the gaze target in the entire dataset, the camera setup in an example scene, and the number of cameras that the gaze target appears.} 
    \label{fig:dataset}
\end{figure*}

\myheading{Multi-view} settings have been extensively studied in tasks such as 3D human pose estimation \cite{ iskakov2019learnable, qiu2019cross, he2020epipolar, dong2019fast, tu2020voxelpose} and 3D reconstruction \cite{furukawa2009accurate, kar2017learning, yao2018mvsnet, chen2021mvsnerf, mildenhall2021nerf, wang2024dust3r}. These methods estimate the 3D locations of human body/hand keypoints or object/scene point clouds and generally require large overlap between different views. In the gaze domain, a few recent works \cite{cheng2023dvgaze, bao2024unsupervised, hisadome2024rotation} have investigated improving gaze direction estimations using multi-view input on specialized datasets \cite{ethxgaze, evedataset}. However, these methods impose several restrictions on the input data: the subject's head cannot turn away from the camera, the face must be rectified, and the eyes should be clearly visible. These limitations prevent these methods from being directly applicable to GTE. \citet{nonaka2022dynamic} introduced a multi-view dataset for 3D gaze estimation, but the subjects wore intrusive eye-tracking glasses, and their proposed model does not consider interactions between views. To our knowledge, no previous work has investigated multi-view GTE, and no dataset is available for training and evaluating this task.

\section{Multi-View Gaze Target (MVGT) Dataset}


For the development and evaluation of multi-view GTE models, we collected a dataset named MVGT. The dataset contains 13,686 images of size 4000$\times$3000 with 2,281 unique gaze targets, resulting in 68,430 pairs when pairing camera views for multi-view GTE. The images were captured by 6 synchronized GoPro Hero 8 cameras from four scenes: a university commons room, a small convenience store, a kitchen, and a research lab space. Images are distributed approximately evenly across scenes. The cameras were controlled by a cellphone via Bluetooth for simultaneous image capturing, and were calibrated before data collection (see Supplementary). \Fref{fig:dataset} shows examples of images and annotations. In total, 28 subjects participated in the data collection, with 7 subjects in each scene. We provide detailed annotations per image, including the subject’s head bounding box (detected with a YOLOv5 head detector \cite{ultralytics2021yolov5, smartconstruction}), the 2D gaze‑target coordinate, and a visibility label. Human annotators mark the laser point as gaze target location and classify each as “target inside,” “target outside,” or “target occluded,” representing 44.3\%, 47.5\%, and 8.2\% of the dataset, respectively. \Fref{fig:dataset}(b) shows most targets are simultaneously visible in at least two camera views. \\
\indent We also introduce a non-intrusive data collection protocol for obtaining precise gaze targets without artifacts. During collection, each subject was instructed to point to a random gaze target with a handheld laser pointer, then turn off the pointer while maintaining their gaze on the target. By comparing images with and without the laser point (\Fref{fig:groundtruth_obtain}), we accurately determined the gaze target location. This approach is both cost-effective and more precise than letting annotators subjectively infer the targets \cite{NIPS2015_ec895663, chong2020detecting}. The dataset only contains images without laser points, while images with laser points were used solely to establish the ground truth. This protocol avoids artifacts on the gaze object compared to using an image-inpainting model to remove the laser point~\cite{gf3d_cvpr23}. Meanwhile, it is easily applicable to new scenes and allows for future extension of the dataset. To introduce pose variability, subjects were asked to stand for half of the samples and sit for the other half. 
\def\subFigSz{0.44\linewidth} 
\begin{figure}[h]
    \centering
    \includegraphics[width=\subFigSz]{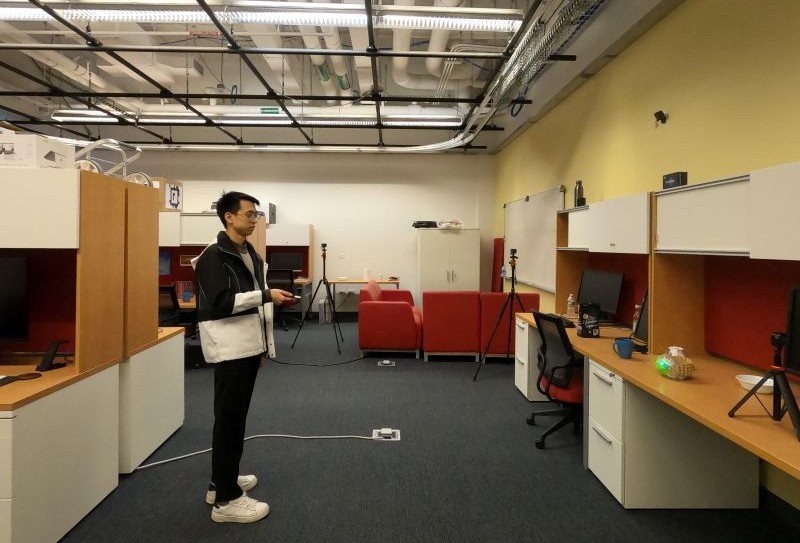} \hfill 
    \includegraphics[width=\subFigSz]{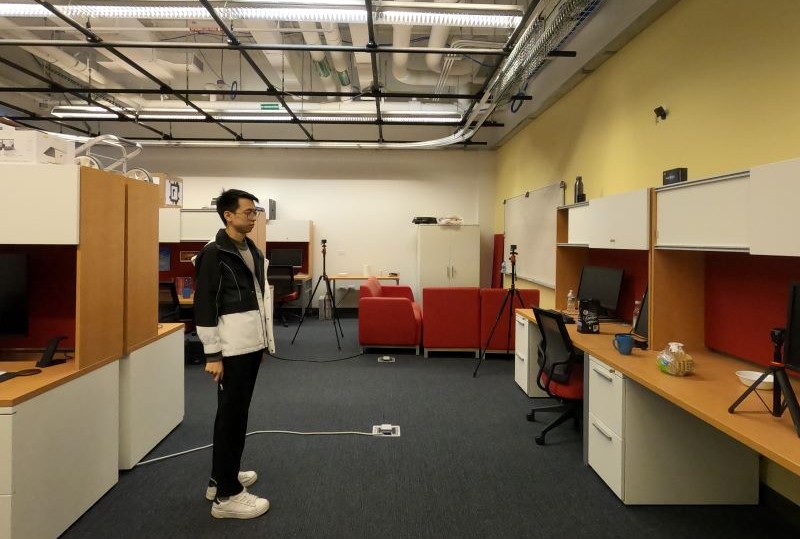} \\
\makebox[\subFigSz]{\small{(a) taken with laser pointer on }} \hfill 
\makebox[\subFigSz]{\small{(b)  taken with laser pointer off}}     
\vskip -0.1in
    \caption{Images of a subject looking at the same gaze target, one with the laser pointer on (a) and one with the laser pointer off (b). }
    \label{fig:groundtruth_obtain}
\end{figure}



\begin{figure*}[t]
\centering
\includegraphics[width=0.95\linewidth]{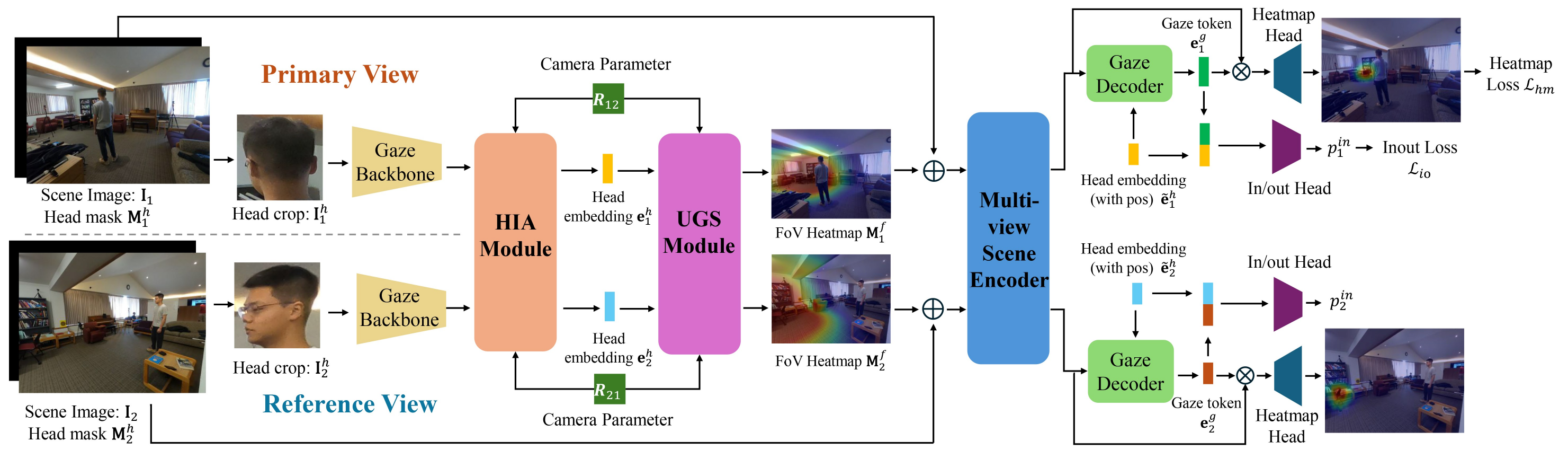} \\
\caption{Overall framework. Our method takes images from a pair of camera views as input. The head images are processed by the HIA and UGS module to output enhanced head embeddings and generate FoV heatmaps from more accurately estimated gaze vectors benefiting from multi-view input. {Camera parameters are provided as input to HIA and UGS to encode the geometric relationship and transform gaze vectors between views.} The FoV heatmaps are input as priors to the multi-view scene encoder with the scene images. The output scene features and head embeddings are fed to a gaze decoder followed by output heads to output the gaze target heatmap and in/out probabilities.} 
\label{fig:framework} \vspace{-0.2cm}
\end{figure*}

\section{Multi-view Gaze Target Estimation}
In this section, we describe our method for multi-view GTE. 
To maximize applicability, our model processes a pair of images as its basic operation, with the potential to analyze more images by aggregating results from multiple pairs. For a pair with a primary view and a reference view, the method predicts the gaze target location and in/out probabilities for each view by leveraging information from the other view.


\subsection{Processing pipeline}

The processing pipeline of our framework is shown in \Fref{fig:framework}. The input consists of a pair of images from the primary and reference views, $\mathbf{I}_1, \mathbf{I}_2 \in \mathbb{R}^{3 \times H \times W}$, and the head bounding boxes of the subject in each view $\vect{x}^{box}_1, \vect{x}^{box}_2  \in \mathbb{R}^4$. We assume that the camera intrinsic parameters, $\mathbf{K}_1, \mathbf{K}_2 \in \mathbb{R}^{3 \times 3}$, and extrinsic parameters, $\mathbf{R}_1, \mathbf{R}_2 \in \mathbb{R}^{3 \times 3}$ and $\mathbf{t}_1, \mathbf{t}_2 \in \mathbb{R}^{1 \times 3}$, are also known for the two views.


Since both images undergo similar processing steps, we will omit the view index for brevity unless specified otherwise. Given an image and the subject's head box, we crop out the head image $\mathbf{I}^h$, also creating a binary mask $\M^h \in \mathbb{R}^{1 \times H \times W}$ of the subject head location in the image. The head image is first processed by the Head Information Aggregation (HIA) module, which outputs the head embedding $\mathbf{e}^h \in \mathbb{R}^d$ after interactions across views. The head embedding is then processed by the Uncertainty-based Gaze Selection (UGS) module to output a more accurate 3D gaze vector $\mathbf{g}$ benefiting from multiple views. Field-of-view (FoV) heatmaps $\M^f \in \mathbb{R}^{1 \times H \times W}$ are computed from the predicted gaze vectors, serving as gaze target priors. These heatmaps are concatenated with the scene images and head masks for each view and are fed into a multi-view scene encoder that contains two Epipolar-based Scene Attention (ESA) modules, outputting the scene feature $\F^s \in \mathbb{R}^{c \times h \times w}$. The scene features and head embeddings are then input to a gaze decoder and the output heads, which produce the gaze target heatmap $\H \in \mathbb{R}^{1 \times 64 \times 64}$ and the probability $p^{in}$ that the target is located in the image.

\begin{figure}[t]
    \centering
    \includegraphics[width=\linewidth]{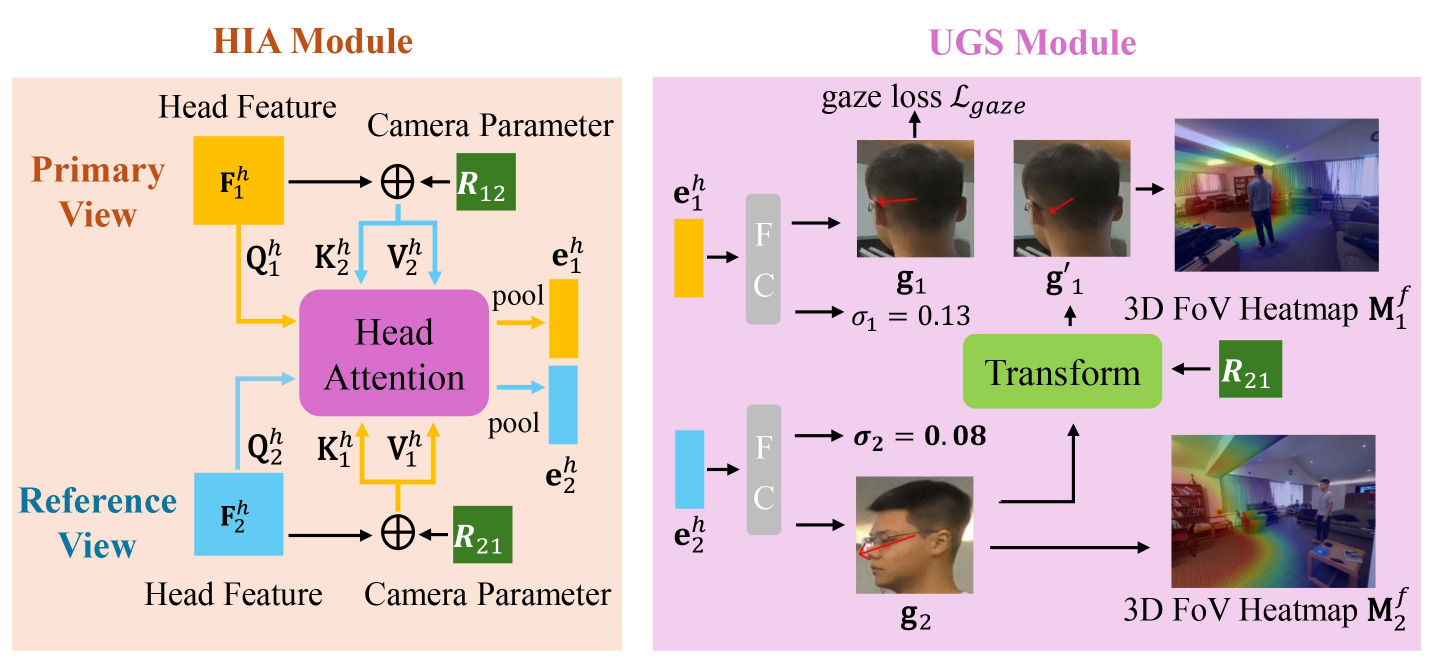}
    \vspace{-0.6cm}
    \caption{Structures of HIA and UGS. HIA aggregates head information from the other view using the head appearances and geometric relationships between views. UGS selects the more reliably predicted gaze vector based on the output uncertainty scores and transforms it to the other view using the camera parameters.}
    \label{fig:gaze_modules}
\end{figure}

\subsection{Head Information Aggregation Module}


We propose the HIA module that leverages head images from both camera views to enhance the head embeddings and improve gaze vector estimations. First, the head feature $\mathbf{F}^h \in \mathbb{R}^{c_0 \times h_0 \times w_0}$ is extracted from the head image $\bm{I}^h$ using a ResNet-18 backbone. Then, $\mathbf{F}^h$ is flattened into tokens of $\mathbb{R}^{c_0 \times h_0 w_0}$. The tokens enter the Head Attention module as queries, which is a cross-attention block to aggregate information from the keys and values of the other view. Take the primary view as an example:
\begin{equation}
\tilde{\mathbf{F}}^{h}_1 = \mathbf{F}^h_1 + CrossAtt(\mathbf{Q}^h_1, \mathbf{K}^h_1, \mathbf{V}^h_1), \\
\end{equation}
where $\mathbf{Q}^h_1 = W^h_q(\mathbf{F}^h_1), \mathbf{K}^h_1 = W^h_k(\mathbf{F}^h_2 \oplus \mathbf{R}_{21})$, $\mathbf{V}^h_1 = W^h_v(\mathbf{F}^h_2 \oplus \mathbf{R}_{21})$, and $W^h_q, W^h_k, W^h_v$ are the linear projection layers. Following \cite{liu2023zero}, we concatenate the keys and values with the relative camera rotation $\mathbf{R}_{21}$ to incorporate the geometric relationship information between views, where $\mathbf{R}_{21} = \mathbf{R}_1 \mathbf{R}_2^{-1}$. As shown in the Supplementary, both the relative rotation and the head appearance from the other view are vital for performance improvement. $\tilde{\mathbf{F}}^{h}_1$ are average-pooled to produce the head embedding $\mathbf{e}^h_1$. The reference view undergoes a similar process to yield $\mathbf{e}^h_2$.

\subsection{Uncertainty-based Gaze Selection Module} \label{sec:ugs}
Although the HIA module helped the information propagation, the two input views can still predict gaze vectors of different qualities. Therefore, we propose the UGS module which picks the more reliably predicted gaze vector from the two views to generate better-quality FoV heatmaps. To find out the more reliable view, we extend the gaze estimator in GTE models to predict an uncertainty score $\sigma$ along with the gaze vector $\g$. We make the model learn the \textit{Aleatoric Uncertainty} \cite{kendall2017uncertainties} of the input image with the uncertainty-aware loss $\mL_{gaze}$ between the predicted gaze vector $\g$ and the ground-truth $\hat{\g}$, similar to \cite{gaze_uncertainty_wacv20}:
\vspace{-4pt}
\begin{equation}
\mathcal{L}_{gaze}(\g, \hat{\g}) =  \frac{1}{2\sigma^2}(1 - \frac{\g  \cdot \hat{\g}}{||\g||_2 ||\hat{\g}||_2}) + \frac{1}{2} log(\sigma^2), \label{eq:gaze_loss}
\end{equation}
 where the first term is the cosine angular loss suppressed by the uncertainty score, and the second is a regularization term. 
 The ground truth gaze vector $\hat{\g }$ can be obtained from the pseudo point cloud using the ground truth gaze coordinate, which we will show later. As shown in previous works~\cite{kendall2017uncertainties, gaze_uncertainty_wacv20}, the model trained with this form of uncertainty loss tends to predict a larger $\sigma$ for the predictions with larger errors.  Therefore, we select the view predicted with lower $\sigma$ and replace the other view's prediction with the one with lower uncertainty using camera transformation: 
 \begin{equation}
     \g_j' = \R_j\R_i^{-1}\g_i, \quad i,j \in \{1, 2\}, \sigma_i < \sigma_j
 \end{equation}
 The gaze vectors are used to generate the FoV heatmaps with the monocular depth maps $\bm{D}$.  The camera intrinsic matrix $\K$ is represented as: 
 \vspace{-4pt}
\begin{equation}
\K  = 
 \begin{bmatrix}
    f^x & 0 & c^x\\
    0 & f^y & c^y\\
    0 & 0 & 1
 \end{bmatrix}. 
 \end{equation}
 Given a pixel coordinate $(u, v)$ in the image, the 3D point cloud $\bm{P}^{(u,v)}=[P^x, P^y, P^z]$ in the camera coordinate system of the input view is represented as: 
 \vspace{-0.2cm}
\begin{equation}
\begin{split}
  &P^x = (u-c^x)/f^x * \bm{D}(u,v), \\
  &P^y = (v-c^y)/f^y * \bm{D}(u,v), \\ 
  &P^z = \bm{D}(u,v).
\end{split}
\label{eq:3d_coordinates}
\end{equation}

 Therefore, the 3D vector from any pixel $(u,v)$ to the subject's eye $(e_x, e_y)$ can be obtained as $\V^{(u,v)} = \bm{P}^{(u,v)} - \bm{P}^{(e_x,e_y)}$, and the ground truth gaze vector $\hat{\g}$ is $\V^{(gt_{x},gt_{y})}$. Similar to \cite{tafasca2023childplay}, they are computed from ``pseudo'' point clouds, as $\bm{D}$ contains relative depth values from a monocular depth estimation model, which is up to a scale and shift factor to the absolute depth $\bm{D}^*$. However, when we use a depth estimation model that has low depth distortion and shift \cite{leres, yin2023metric3d}, we can ignore the shift term, and $\V^{(u,v)}$ will only be the same due to the elimination of the scale factor based on Eq.~(\ref{eq:3d_coordinates}). 
 Based on the pseudo point clouds, we obtain the value at $(u,v)$ in the FoV heatmap:

\begin{equation}
  \M^f(u,v) = max (0, \frac{\V^{(u,v)} \cdot \g}{||\V^{(u,v)}||_2 ||\g||_2}).  \label{eq:fov_compute}
\end{equation}

We adopt an exponential decay scheme if the value on the FoV heatmap is lower than 0.9, as in \cite{tafasca2023childplay}. As shown in Supplementary, we observe larger $\sigma$ values for gaze vectors with larger errors, and by selecting the more reliable view, UGS improves the FoV heatmap and final prediction, especially when the original predicted vector has a large error.


\begin{figure}[t]
    \centering
    \includegraphics[width=\linewidth]{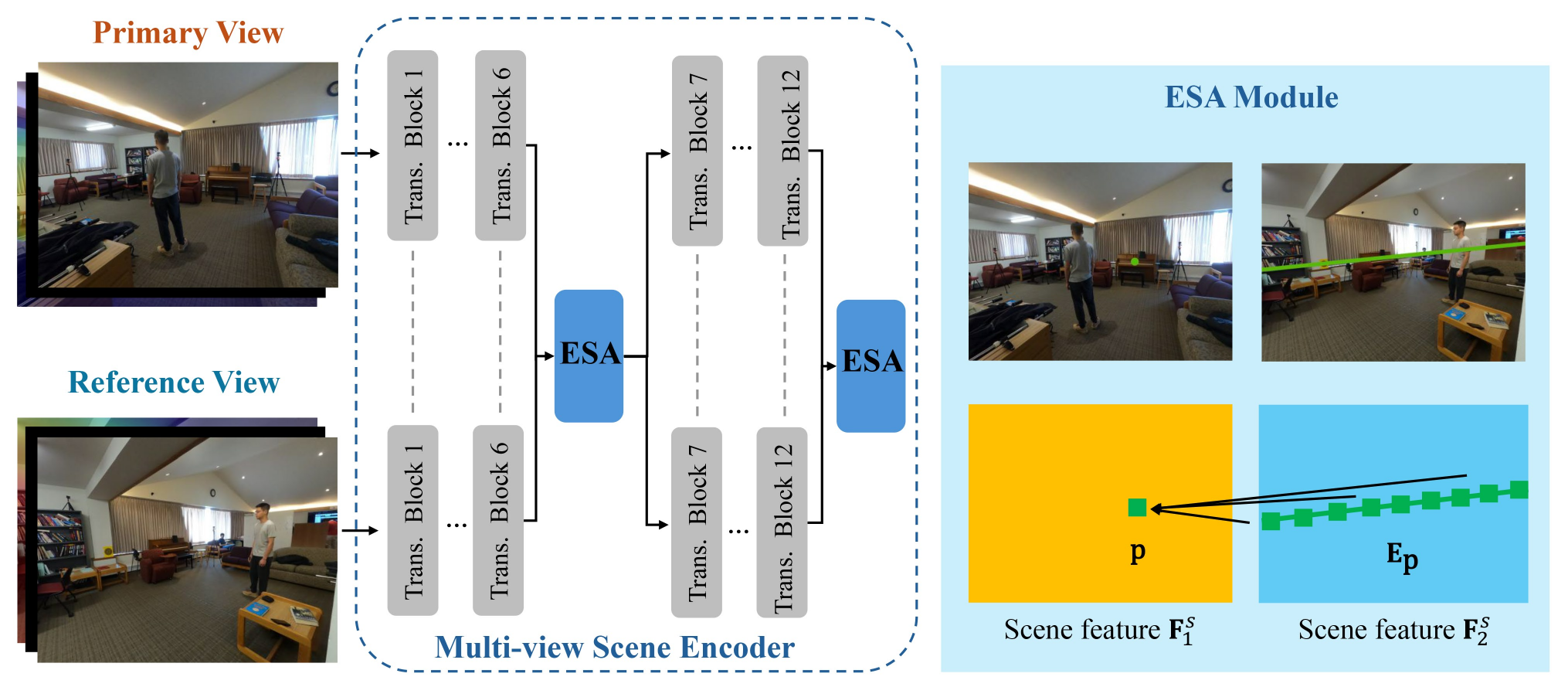}
    \caption{Structure of the multi-view scene encoder and ESA module. The transformer blocks are shared between the two input views. Two ESA modules are inserted in the transformer encoder. In ESA, each feature token attends to multiple tokens sampled along the epipolar line in the other view.}
    \vspace{-0.2cm}
    \label{fig:scene_encoder}
\end{figure}

\subsection{Multi-view Scene Encoder}
The FoV heatmap $\M^f$ is used as priors and concatenated with the scene image $\mathbf{I}$ and the head mask $\M^h$ for both views, and input to the multi-view scene encoder. As shown in \Fref{fig:scene_encoder}, the multi-view scene encoder consists of a ViT-base \cite{vit} encoder along with two Epipolar-based Scene Attention~(ESA) modules for propagating scene information between views. The transformer blocks are shared between
views. In the ESA module, each token in one view attends to the feature tokens uniformly sampled along the epipolar line of the other view. Take a token $\p$ with a coordinate of $(u,v)$ in the primary view feature $\F^s_1$ as an example:
\begin{equation}
    \p' = \p  + CrossAtt(W^s_q(\p ), W^s_k(\E_{\p }),  W^s_v(\E_{\p})),
\end{equation}
where $\E_{\p} = \{\q_{l}\}_{l=1}^{N}$ are the feature vectors sampled along the epipolar line of $\p $ in $\F^s_2$. The epipolar line is computed from the fundamental matrix: $l^{epi}=\mathbb{F}\vect{x}$, where  $\vect{x}=[u, v, 1]^T$ and $\mathbb{F}$ is computed from the camera parameters using multi-view geometry~\cite{multiview_geometry}. Epipolar attention has been used in multi-view tasks including 3D reconstruction~\cite{yang2022mvs2d, huang2024epidiff} and pose estimation~\cite{he2020epipolar}, where it enhances the feature with another view's appearance associated in 3D, especially in the case of occlusion in the primary view. In our case, it also saves computation and memory compared to dense cross-attention on the higher-resolution scene features. We observed that ESA improves GTE performance when the other view contains the gaze target.

\subsection{Output and Losses}
In the final stage, the head embedding $\Tilde{\vect{e}}^h$ and the scene feature $\F^s$ from the scene encoder is fed to a gaze decoder. The head embedding $\Tilde{\vect{e}}^h$ is $\vect{e}^h$ from the HIA module added with a positional encoding $\vect{e}^{pos}$, which is mapped from the normalized head center coordinate with an MLP.  The head embedding $\Tilde{\vect{e}}^h$ is used as a query while $\F^s$ serves as the key and value.  The gaze decoder outputs a gaze token $\vect{e}^g$. For gaze target estimation,  $\vect{e}^g$ is element-wise multiplied with each token in $\F^s$, and fed into a heatmap head to output the gaze target heatmap $\H$. On the other hand, the head embedding $\Tilde{\vect{e}}^h$ is also concatenated with the gaze embedding $\vect{e}^g$ to output the probability of the target located in the frame $p^{in}$.

The overall training loss is formulated as:
\begin{equation}
    \mathcal{L} = \alpha  \mathcal{L}_{hm} + \beta  \mathcal{L}_{io} + \lambda  \mathcal{L}_{gaze},  \label{eq:finalloss}
\end{equation}
where $\mL_{hm}$ is the Mean-Square-Error (MSE) loss between the predicted heatmap $\H$ and the ground truth heatmap $\hat{\H}$ which is a Gaussian centered at the ground truth gaze target coordinate.  $\mL_{io}$ is a binary cross-entropy loss between $p^{in}$ and the ground truth in/out label. For gaze targets labeled with ``occlusion'', we assign them to the ``inside'' class for the in/out task, as these targets remain within the frame but are merely obscured by other objects.

\subsection{Cross-view Estimation}
\begin{figure}[t]
    \centering
    \includegraphics[width=\linewidth]{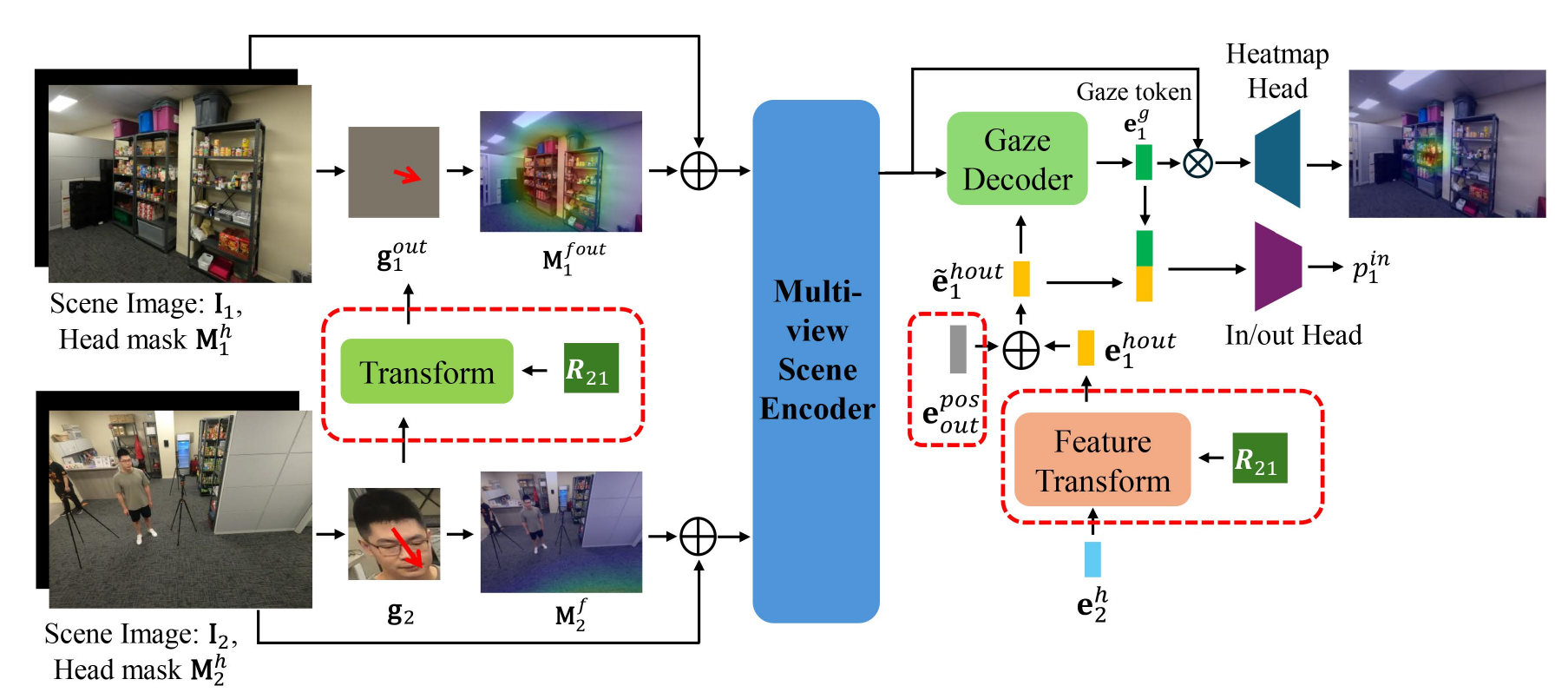}
    \vspace{-0.5cm}
    \caption{The modified model structure for cross-view GTE. The parts encircled in red are the modules we modified/added for the cross-view gaze estimation cases. Modules for estimating the gaze vectors are the same as above and omitted.}
    \label{fig:cross_view}
\end{figure}


Our method can be extended to address cross-view estimation, where the primary view only contains the gaze target, and the subject is only visible in the reference view. In this case, we need to generate the FoV heatmap $\M^f_1$ using the eye location $\bm{P}_2^{(e_x,e_y)}$ and the predicted gaze vector $\g_2$ from the reference view. However, as mentioned in \Sref{sec:ugs}, the point clouds were ``pseudo'' point clouds of which the depth values are up to a scale and shift to the real absolute depth, so the actual 3D location of $\bm{P}_2^{*(e_x,e_y)}$ is not known. Therefore, the eye location $\bm{P}_1^{*e}$ in the primary view's camera coordinate system cannot be directly obtained. 

We use a Multi-view Stereo model to reconstruct the scene in 3D to obtain the absolute depth values. As in the cross-view cases, the input pair of view usually has little or no overlap (\Fref{fig:cross_view}), we assume that a set of images from all six cameras have been obtained before the capturing of input data, so that the 3D scene can be reconstructed. We use, e.g., six images collected in extrinsic parameters calibration to reconstruct the 3D scene with a SOTA multi-view reconstruction model, Dust3R \cite{wang2024dust3r}. When inputting the camera parameters calibrated in real-world metrics, Dust3R can generate depth estimations that are very close to the absolute depth values by optimizing a reconstruction loss \cite{wang2024dust3r}. After obtaining the absolute depth values for both views, for each image, we estimate the scale and shift between the monocular depth map in each new input image and the absolute depth in the reconstructed scene. In this way, $\bm{P}_2^{*(e_x,e_y)}$ can be obtained, along with $\bm{P}_1^{*e}$ via camera transformation, from which $\M^f_1$ can be generated. We provided the detailed procedure for scale and shift estimation and some reconstruction examples in Supplementary.


On the other hand, we also updated the later part of the model for cross-view GTE (\Fref{fig:cross_view}). As the primary view does not contain the subject's head, we add a feature transform module to generate the head embedding $\vect{e}^{hout}_{1}$ from the reference view. The feature transform module is a two-layer MLP, which takes the concatenated input of the reference view's head embedding $\vect{e}^{h}_{2}$ and the relative camera rotation $R_{21}$. Meanwhile, we add a learnable ``outside embedding" $\vect{e}^{pos}_{out}$ to $\vect{e}^{hout}_{1}$ as the positional embedding to get the final head embedding for primary view. In the experiments, we fine-tuned our model trained for the ordinary multi-view setting above on the view pairs that fall in the cross-view GTE category. As will be seen, our method can predict the cross-view gaze targets well.


\section{Experiments}

\subsection{Experimental setups}

\begin{figure*}[t]
\begin{floatrow}
\centering
\footnotesize
\capbtabbox[0.65\textwidth]{
\begin{tabular}{l|cccc|cccc}
\Xhline{2\arrayrulewidth}
\multirow{3}{*}{Method} & \multicolumn{4}{c|}{Head Visible}                                                            & \multicolumn{4}{c}{Head Not Visible}                                                            \\ 
  & \multicolumn{2}{c}{Target Visible}                            & \multicolumn{2}{c|}{Target Not Visible}      & \multicolumn{2}{c}{Target Visible}                            & \multicolumn{2}{c}{Target Not Visible}       \\ \cline{2-9} 
     & \multicolumn{1}{c}{Dist. $\downarrow$} & \multicolumn{1}{c|}{AP $\uparrow$} & \multicolumn{1}{c}{Dist. $\downarrow$} & AP $\uparrow$ & \multicolumn{1}{c}{Dist. $\downarrow$} & \multicolumn{1}{c|}{AP $\uparrow$} & \multicolumn{1}{c}{Dist. $\downarrow$} & AP $\uparrow$ \\ \Xhline{2\arrayrulewidth}
Random   & 0.451   & \multicolumn{1}{c|}{0.555}   &   0.456  &  0.546   & 0.464     & \multicolumn{1}{c|}{0.502}   & \multicolumn{1}{c}{0.457}     &  0.564  \\ 
Center   & 0.259   & \multicolumn{1}{c|}{/}   &   0.261  &  /   & 0.272     & \multicolumn{1}{c|}{/}   & 0.253     &  /  \\ 
Chong \cite{chong2020detecting}    &    0.159                      & \multicolumn{1}{l|}{0.855}   &         0.157                 &  0.862  &  0.191                        & \multicolumn{1}{l|}{0.792}   &     0.174                     &  0.866   \\
Miao \cite{miao2023patch}    &     \underline{0.141}                      & \multicolumn{1}{c|}{0.886}   &         \underline{0.140}                 &  0.892  &  \underline{0.164}                        & \multicolumn{1}{l|}{\underline{0.831}}   &     0.163                     &  \textbf{0.886}   \\
Tafasca* \cite{tafasca2023childplay}  &      0.149                    & \multicolumn{1}{c|}{\underline{0.893}}   &   0.148                       & \underline{0.895}    &  0.166                        & \multicolumn{1}{c|}{\underline{0.831}}   &         \underline{0.154}                 &  \underline{0.884}  \\
Ours-Single &   0.151                       & \multicolumn{1}{c|}{0.877}   &      0.148                    &  0.878  &      0.179                    & \multicolumn{1}{c|}{0.758}   &         \underline{0.154}                &  0.855  \\
Ours     &       \textbf{0.129}                   & \multicolumn{1}{l|}{\textbf{0.909}}   &         \textbf{0.122}                 &   \textbf{0.912}  &        \textbf{0.161}                  & \multicolumn{1}{c|}{\textbf{0.836}}   &         \textbf{0.152}                 &  0.868   \\ \Xhline{2\arrayrulewidth}
\end{tabular}
}
{\caption{Comparing with single-view GTE methods on test data divided based on the head and target visibility in the reference view. Best numbers are marked as bold and 2nd best are underlined. Our method shows large improvement when the subject head is visible, while maintaining comparable performance when the head is not visible.} \label{tab:main_results}}
\hfill
\ffigbox[0.34\textwidth]{
    \centering
    \includegraphics[width=\linewidth]{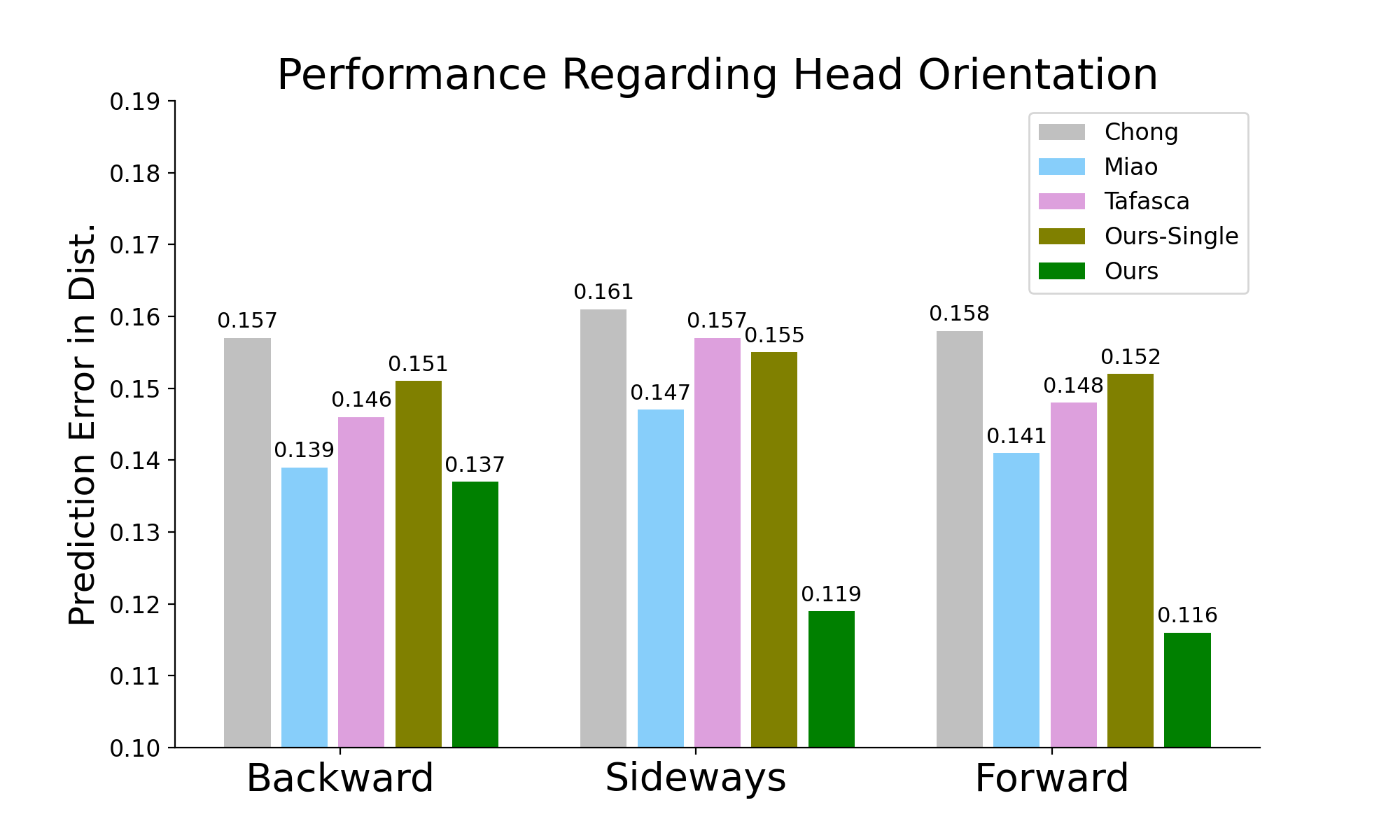}
    }
    {\vspace{-0.5cm}
    \caption{Results regarding different head orientations in the reference view. Our method shows a much larger improvement when the subject face is half/fully visible. \label{fig:head_orient_analysis}  }}  
\end{floatrow} \vspace{-5pt}
\end{figure*}

\myheading{Implementation details.} We process the scene image at a resolution of $512 {\times} 384$ and the head crop at a resolution of $224 {\times} 224$. The gaze backbone is a ResNet-18 \cite{resnet} pretrained on Gaze360 \cite{gaze360}, while the transformer part of the multi-view scene encoder is a ViT-base \cite{vit} model pretrained with MultiMAE \cite{multimae}. We used Metric3D~\cite{yin2023metric3d} for monocular depth estimation. In ESA, we sampled 48 feature vectors along the epipolar line for each query token.

\myheading{Training and evaluation.} To simulate the real-world application of applying a trained model to a new scene with multiple camera setups, we performed leave-one-scene-out cross-validation in our experiments. In the experiments, all models are first trained on the GazeFollow dataset \cite{NIPS2015_ec895663}, fine-tuned on three scenes of our MVGT dataset, and then validated on the left-out scene. Each of the four scenes was left out for validation, and the results averaged across scenes are reported. For our method, we train the single-view version of the model on GazeFollow and fine-tune the whole model on the MVGT dataset. We used a batch size of 40 pairs of views. The specific parameter settings for validating each scene are described in the Supplementary.

\myheading{Evaluation metrics.} We use the normalized $L_2$ Distance (\textbf{Dist.}) between the predicted gaze target coordinates and the ground truth gaze target annotations to evaluate GTE performance. AUC is not used because it is more suited for evaluating the alignment of predicted heatmaps with group-level annotations, making it less suitable for datasets with single-point annotations, as explained in \cite{tafasca2024sharingan}. We use \textbf{AP} to evaluate the performance on in/out classification.

\subsection{Comparison with Single-View Methods}

We compare our multi-view GTE method with SOTA single-view GTE methods, and the single-view baseline version of our method (\textit{Ours-Single}). In \textit{Ours-Single}, we exclude the HIA, UGS, and ESA modules to eliminate interaction between views, and the gaze estimator only outputs a single gaze vector without the uncertainty score. 
We evaluate the model on primary views, treating the reference view as additional input. To ensure a fair comparison with single-view methods, when a primary view image (e.g., Camera1) is paired with different cameras (Camera2, Camera3, etc.) in multiple pairs, we evaluate on the primary view for each pair and average the scores for the same primary view image. This ensures the same total number of testing samples, enabling direct comparison between methods.


We experimented with the following baseline methods: \textit{Random} generates heatmap response and in/out probability randomly in a [0,1] uniform distribution. \textit{Center} generates a heatmap always at the image center. \textit{Chong} \cite{chong2020detecting} is a popular GTE model that only uses RGB input. \textit{Miao} \cite{miao2023patch} uses monocular depth maps as direct input for GTE. \textit{Tafasca} \cite{tafasca2023childplay} is a recent model that generates a FoV heatmap from a predicted 3D gaze vector and a monocular depth map. We re-implemented the model and were able to reproduce its performance on the GazeFollow dataset (See Supplementary). 

To better understand the benefits of using an additional view, we divide the test cases into four categories based on the visibility of the subject's head and the gaze target in the reference view. The number of samples for each category is shown in Supplementary. In Tab.\ref{tab:main_results}, our method shows significant improvement when the reference view contains the head, while maintaining comparable performance with the best baselines when it does not. The improvement is most pronounced when the reference view includes the head but not the gaze target, which typically occurs when the subject is facing toward the camera with a clear face appearance. 

\begin{figure*}[t]
\begin{floatrow}
\centering
\ffigbox[0.6\textwidth]{
\includegraphics[width=\linewidth]{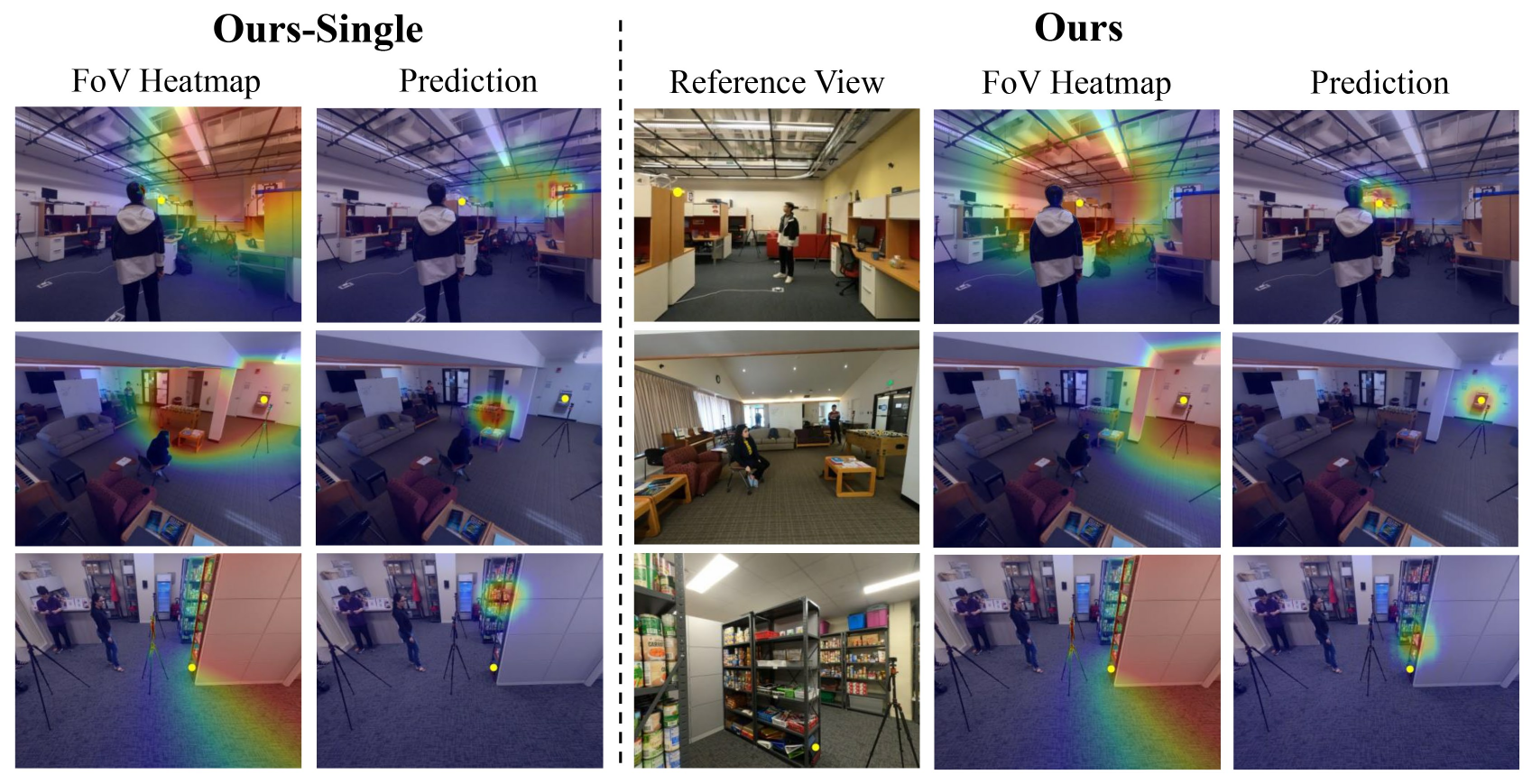} 
}
{   \vspace{-0.6cm}  
    \caption{Qualitative comparisons of our method with and without multi-view processing. Yellow dots indicate ground truth. Our method can leverage the head appearance in the reference view to obtain better FoV heatmaps (Rows 1\&2), and get enhanced performance using the scene appearance from the other view (Row3).}
    \label{fig:qualitative_main}
}
\hfill
\ffigbox[0.36\textwidth]{
\includegraphics[width=\linewidth]{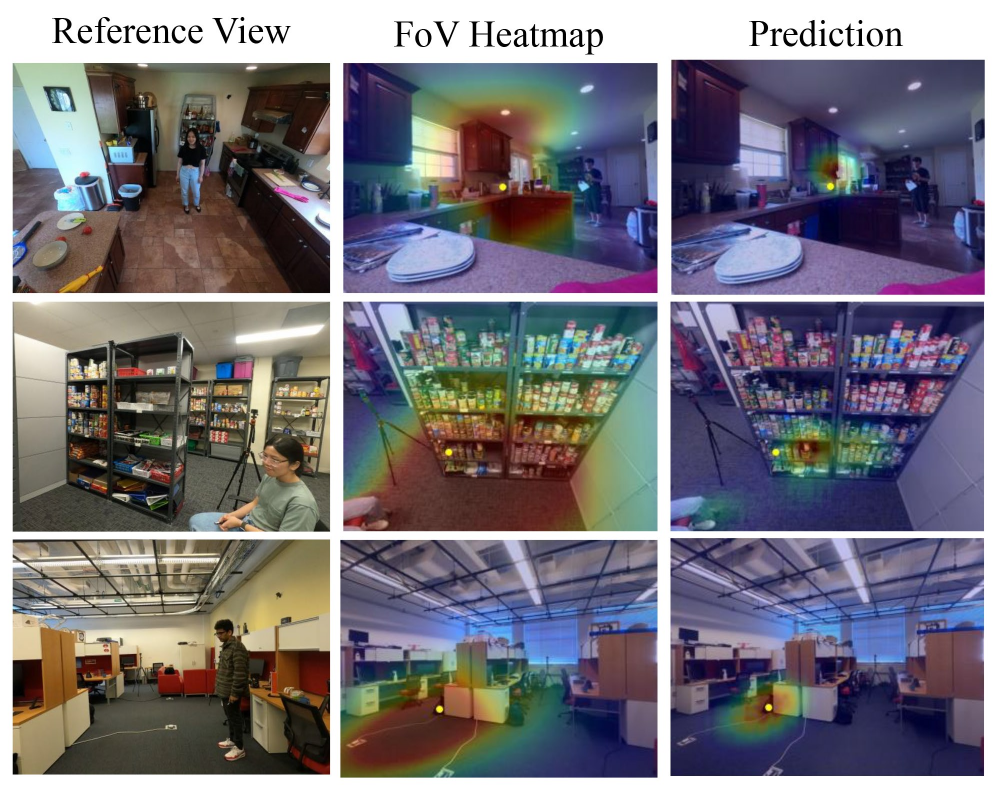}
}
{
\vspace{-0.6cm}  
\caption{Qualitative examples for cross-view GTE. The reference view input, and the primary view's FoV heatmaps and predicted heatmaps are shown. Ground truth is shown as yellow dots.} 
\label{fig:cross_view_results}
}
\end{floatrow}
\end{figure*}

 The benefits of the reference view are further highlighted by comparing to our method without the reference view and multi-view processing (Ours-single). The advantage of the reference view is clear, except when neither the head nor the target is visible in it, as expected, since little additional information can be gained in this case. As shown in \Fref{fig:qualitative_main}, our methods can leverage the useful face information from the reference view and obtain higher-quality FoV heatmaps (Rows 1-2). It also benefits from the appearance of the gazed objects in a different perspective (Row 3). This demonstrates the effectiveness of our method in leveraging two-view input. {Using more than two views can further improve performance, as shown in the Supplementary.}


\myheading{Performance Regarding Head Orientation.} We further investigate the effect of the face visibility when the reference view contains the subject's head, by analyzing the head orientation. We use a head pose estimation model \cite{hopenet} to obtain the yaw angle from the head image. The head pose estimator does not perform well when the head is facing away from the camera, so we combine it with a face keypoint estimator \cite{face_alignment} to divide the head orientations into three categories: backward: $< 30$ face keypoints are detected; sideways: $\geqslant 30$ face keypoints are detected and the head pose yaw angle $\geqslant$ 55\textdegree; forward: the remaining images.

The results are shown in \Fref{fig:head_orient_analysis}. As illustrated, having the head visible in the reference view consistently provides benefits, reducing the distance error. The error reduction rate is 23.7\% and 23.2\% for forward and sideways orientations, respectively---significantly greater than the 9.3\% error reduction rate when the person is facing backward.


\begin{table*}[h]
\begin{floatrow}
\centering
\setlength{\tabcolsep}{4pt}
\capbtabbox[0.68\textwidth]{
\footnotesize
\begin{tabular}{cccc|cccc|cccc}
\Xhline{2\arrayrulewidth}
\multirow{3}{*}{$\sigma$} & \multirow{3}{*}{HIA} & \multirow{3}{*}{UGS} & \multirow{3}{*}{ESA} & \multicolumn{4}{c|}{Head Visible}                                                            & \multicolumn{4}{c}{Head Not Visible}                                                            \\ 
  &  & & & \multicolumn{2}{c}{Target Visible}                            & \multicolumn{2}{c|}{Target Not Visible}      & \multicolumn{2}{c}{Target Visible}                            & \multicolumn{2}{c}{Target Not Visible}       \\ \cline{5-12} 
    &  & & &  \multicolumn{1}{c}{Dist. $\downarrow$} & \multicolumn{1}{c|}{AP $\uparrow$} & \multicolumn{1}{c}{Dist. $\downarrow$} & AP $\uparrow$ & \multicolumn{1}{c}{Dist. $\downarrow$} & \multicolumn{1}{c|}{AP $\uparrow$} & \multicolumn{1}{c}{Dist. $\downarrow$} & AP $\uparrow$\\ \Xhline{2\arrayrulewidth}
&  & & &  0.151   & \multicolumn{1}{c|}{0.877}   &   0.148  &  0.878   & 0.179     & \multicolumn{1}{l|}{0.758}   & \multicolumn{1}{c}{0.154}     &  0.855  \\ 
\checkmark &  & & &        0.145                  & \multicolumn{1}{l|}{0.874}   &   0.147                       & 0.874    &  0.177                        & \multicolumn{1}{l|}{0.756}   &         \textbf{0.151}                 &  0.855 \\
\checkmark &  \checkmark & & &    0.135                    & \multicolumn{1}{l|}{0.896}   &   0.133                       & 0.897    &  0.174                        & \multicolumn{1}{l|}{\underline{0.821}}   &         0.153                 &  \textbf{0.873}  \\
\checkmark & \checkmark & \checkmark  &   &   \underline{0.130}                         & \multicolumn{1}{l|}{\underline{0.902}}   & \underline{0.123}                &  \underline{0.908}    &   \underline{0.170}                & \multicolumn{1}{l|}{ 0.810 }   &    \underline{0.152}                   & 0.863   \\
\checkmark&  \checkmark & \checkmark & \checkmark &       \textbf{0.129}                   & \multicolumn{1}{l|}{\textbf{0.909}}   &         \textbf{0.122}                 &   \textbf{0.912}  &        \textbf{0.161}                  & \multicolumn{1}{l|}{\textbf{0.836}}   &         \underline{0.152}                 &  \underline{0.868}   \\ \Xhline{2\arrayrulewidth}
\end{tabular}
\vspace{-6pt}
}
{\caption{Ablation Study Results. All three modules demonstrate their effectiveness in leveraging multi-view information to improve GTE accuracy. 
\label{tab:ablation_results} \vspace{-6pt}
}
}
\capbtabbox[0.25\textwidth]{
\footnotesize
\begin{tabular}{l c c}
\Xhline{2\arrayrulewidth}
Method & Dist. $\downarrow$  & AP $\uparrow$ \\
\Xhline{\arrayrulewidth}
Random &  0.446   & 0.462  \\
Center &  0.245   &  /  \\
DeepGazeIIE \cite{deepgazeiie} &  0.248 & / \\
Recasens \cite{recasens2017following} & 0.271 & 0.542 \\
Ours  & \textbf{0.188}  & \textbf{0.820} \\
\Xhline{2\arrayrulewidth}
\end{tabular}
}
{
\caption{Cross-view GTE results.}
\label{tab:cross_view_results}
}
\end{floatrow} \vspace{-10pt}
\end{table*}

\subsection{Ablation Study}
Tab. \ref{tab:ablation_results} shows the ablation study results. The first row shows the single-view baseline version of our method. When the gaze estimator is extended to predict an uncertainty score~$\sigma$, the model shows a small improvement due to better gaze vector prediction \cite{gaze_uncertainty_wacv20}. The HIA module significantly improves the performance by incorporating head information from the reference view and the geometry relations, which lead to more accurate gaze vectors and enhanced head embeddings when input to the output heads. The UGS module further improves the performance by selecting the more accurate gaze vector in the two views. The ESA modules provide further improvement when the gaze target is visible (1st and 3rd columns), suggesting that they add helpful scene context for GTE.  The improvement is relatively less when the head is visible in reference because the gaze vectors and the corresponding FoV heatmaps determine the priors for the potential attended area, instead of the scene background. {See the Supplementary for more detailed analyses of the proposed modules. Notably, the benefits of incorporating camera parameters as input in the HIA module for the in/out prediction task, are evident.}

\subsection{Cross-View Estimation}
In this section, we train the model to predict the gaze target in the primary view which contains the target but not the subject, by using the person's appearance in the reference view. We fine-tune the model trained in the ordinary multi-view setting above on these cross-view camera pairs (around 7000 pairs).
In this case, none of the single-view GTE models can predict the targets or serve as baseline methods. {The strongest baseline we propose is adapting a method for predicting gaze targets in future video frames from the current observed frame \cite{recasens2017following} (outside of the current observed frame). We fine-tune it on the cross-view samples in our dataset, treating the reference view as the ``current frame'' and the primary view as the ``future frame.'' We also evaluated DeepGazeIIE \cite{deepgazeiie} as a representative baseline of the saliency prediction model.} 
\Tref{tab:cross_view_results} shows that our method outperforms the other approaches by a wide margin. The qualitative examples in \Fref{fig:cross_view_results} demonstrate that our method generates reasonable FoV heatmaps in the primary view based on the person's appearance in the reference view, and predict target locations reasonably well.

\section{Conclusions}

This paper proposed the first method for multi-view gaze target estimation (GTE). The model incorporates a Head Information Aggregation (HIA) module to aggregate head information, an Uncertainty-based Gaze Selection (UGS) module to select the more reliable gaze vector predicted, and Epipolar-based Scene Attention (ESA) module for integrating scene background information. Our method shows large improvements when the reference view contains the person's head, and can be extended to cross-view GTE that single-view methods cannot handle. In addition, we introduced the MVGT dataset, the first dataset for multi-view GTE with calibrated camera parameters and precisely annotated targets. Future work could explore learning geometric-aware features without inputting camera parameters, or address the cross-view task without access to the reconstructed 3D scene. We expect our work can draw more attention to using multi-view input in the GTE domain.

\myheading{Acknowledgments.} This project was partially supported by NSF award IIS-2123920 and the Department of Surgery at Stony Brook University. 
Minh Hoai was initially supported by NSF award DUE-2055406, and later in part by the Australian Institute for Machine Learning (University of Adelaide) and the Centre for Augmented Reasoning, an initiative of the Australian Government's Department of Education.
The authors thank Haoyu Wu for helpful discussions.

{
    \small
    \bibliographystyle{ieeenat_fullname}
    \bibliography{main}
}
\newcolumntype{?}{!{\vrule width 0.8pt}}

\setcounter{section}{0}
\setcounter{figure}{0}
\setcounter{table}{0}
\renewcommand\thesection{S\arabic{section}}
\renewcommand\thefigure{S\arabic{figure}}
\renewcommand\thetable{S\arabic{table}}

\clearpage
\setcounter{page}{1}
\maketitlesupplementary

\begin{abstract}
In this supplementary material, we provide additional information for the implementation details (\ref{sec:imp_details}) and statistics of the MVGT dataset (\ref{sec:dataset_stats}). We extend our model to use more than 2 views (\ref{sec:view_extend}). We provide detailed information for the absolute depth computation (\ref{sec:absolute_depth}) and camera calibration procedure (\ref{sec:calibration}). We show the angular errors of the predicted gaze vectors of our model (\ref{sec:angular_error}). We provide further analyses of the HIA (\ref{sec:hia_analysis}), UGS (\ref{sec:ugs_analysis}), and ESA (\ref{sec:esa_analysis}) modules. We analyze the sensitivity of the model to the camera parameters (\ref{sec:camera_sensitivity}) and analyzed the complexity and costs of our model (\ref{sec:modelcost}). We show the performance of training our model with the same learning rate when evaluating on different scenes (\ref{sec:samelr}).  We show the performance of a reproduced baseline model on the GazeFollow dataset (\ref{sec:reproduce}). We provide additional discussions of the applicability, generalization, and limitations of our dataset and model (\ref{sec:discussions}). 

\end{abstract}

\section{Implementation Details} \label{sec:imp_details}
In the overall loss computation, we used $\alpha=10.0$ and $\lambda=0.1$. As mentioned in the main text, we performed leave-one-scene-out cross-validation in our experiments. $\beta$ was set as 0.05 when leaving out the commons room scene for evaluation, and 0.3 when evaluating other scenes. We used a learning rate of $5 \times 10^{-5}$ and a batch size of 40 when pretraining the single-view version of our model on GazeFollow \cite{NIPS2015_ec895663}. When fine-tuning the model on MVGT, we first train the gaze estimator with a learning rate of $1 \times 10^{-4}$ with a batch size of 60. Then we fine-tune the full model using a batch size of 40 pairs of views. The learning rate is $2.5 \times 10^{-6}$ when evaluating the lab and store scenes, $2.5 \times 10^{-5}$ when evaluating the commons scene, and $1 \times 10^{-5}$ for the kitchen scene. For the cross-view task, due to the relatively small number of samples, we used a learning rate $1 \times 10^{-7}$ for the research lab and commons room scenes, and $1 \times 10^{-8}$ for the store and kitchen scenes . To obtain the eye location of the subject when computing the field-of-view (FoV) heatmap, we apply a face keypoint estimator \cite{face_alignment} to the head crop of the subject and locate the eye keypoint location. If no eye is detected (e.g., the person is looking away from the camera), we choose the center point of the head bounding box as the eye location.

\section{Additional Dataset Information} \label{sec:dataset_stats}

\begin{table}[t]
\footnotesize
\begin{NiceTabular}{c|c?cc|cc}
\Xhline{2\arrayrulewidth}
\Block{2-2}
  {
    \diagbox{Primary View}{Reference View}
  }                      &                                                            & \multicolumn{2}{c|}{Head Vis.}                                                                                                           & \multicolumn{2}{c}{Head Not Vis.}                                                                                                        \\ \cline{3-6} 
                              &                                                            & \multicolumn{1}{c|}{\begin{tabular}[c]{@{}c@{}}Target \\ Vis.\end{tabular}} & \begin{tabular}[c]{@{}c@{}}Target \\ Not Vis.\end{tabular} & \multicolumn{1}{c|}{\begin{tabular}[c]{@{}c@{}}Target \\ Vis.\end{tabular}} & \begin{tabular}[c]{@{}c@{}}Target \\ Not Vis.\end{tabular} \\ \Xhline{2\arrayrulewidth}
\multirow{3}{*}{\makecell{Head\\Vis.}} & \makecell{Target\\Vis.}
& \multicolumn{1}{c|}{\cellcolor{green!10!white}11216} & \cellcolor{green!10!white}12314 & \multicolumn{1}{c|}{\cellcolor{green!10!white}660} & \cellcolor{green!10!white}2485 \\ \cline{2-6}

& \makecell{Target\\Not Vis.}
& \multicolumn{1}{c|}{\cellcolor{green!10!white}12314}
& \cellcolor{green!10!white}13164
& \multicolumn{1}{c|}{\cellcolor{green!10!white}1254}
& \cellcolor{green!10!white}2738 \\ \hline

\multirow{3}{*}{\makecell{Head\\ Not Vis.}}     & \makecell{Target\\Vis.}     & \multicolumn{1}{c|}{ \cellcolor{red!10!white} 660}                                                    & \cellcolor{red!10!white}1254                                                       & \multicolumn{1}{c|}{526}                                                    & 1210                                                       \\ \cline{2-6} 
  & \makecell{Target\\Not Vis.} & \multicolumn{1}{c|}{\cellcolor{red!10!white}2485}                                                   & \cellcolor{red!10!white}2738                                                       & \multicolumn{1}{c|}{1210}                                                   & 2202                                                       \\ \Xhline{2\arrayrulewidth}
\end{NiceTabular}
\caption{Statistics of the camera view pairs in MVGT dataset regarding head and face visibilities in both the primary and reference views. Cells shaded in green are the camera view pairs used in comparison with single-view GTE methods. Cells shaded in red correspond to the pairs used in the cross-view GTE experiment.}
\label{tab:dataset_division}
\end{table}

Here we provide more detailed statistics on the division of of the dataset regarding the head and gaze target visibilities in the reference view. As shown in Tab.\ref{tab:dataset_division}, we divide all the 68,430 camera view pairs into $4 \times 4$ cells according to the head and face visibilities of the primary and reference view. In Tab.1 in the main paper, we compared with single-view gaze target estimation (GTE) baselines, and evaluated the models on the primary view images. The cells shaded in green are the samples used in this scenario, as the primary view images must be applicable for GTE  with themselves (head is visible). Within those samples, the ones with the target visible for the primary view are used for the GTE task, while all the samples shaded in green are used for the in/out classification task. On the other hand, the cells shaded in red are the pairs that we used for cross-view GTE and in/out classification, in which the head is only visible in the reference view. The rest of the pairs (white) are not used in any experiments, as neither single-view nor multi-view GTE methods are applicable due to the head not being visible in either view.

\begin{figure*}[t]
    \centering
    \begin{subfigure}[b]{\textwidth}
    \centering
    \includegraphics[width=0.8\linewidth, height=2.7in]{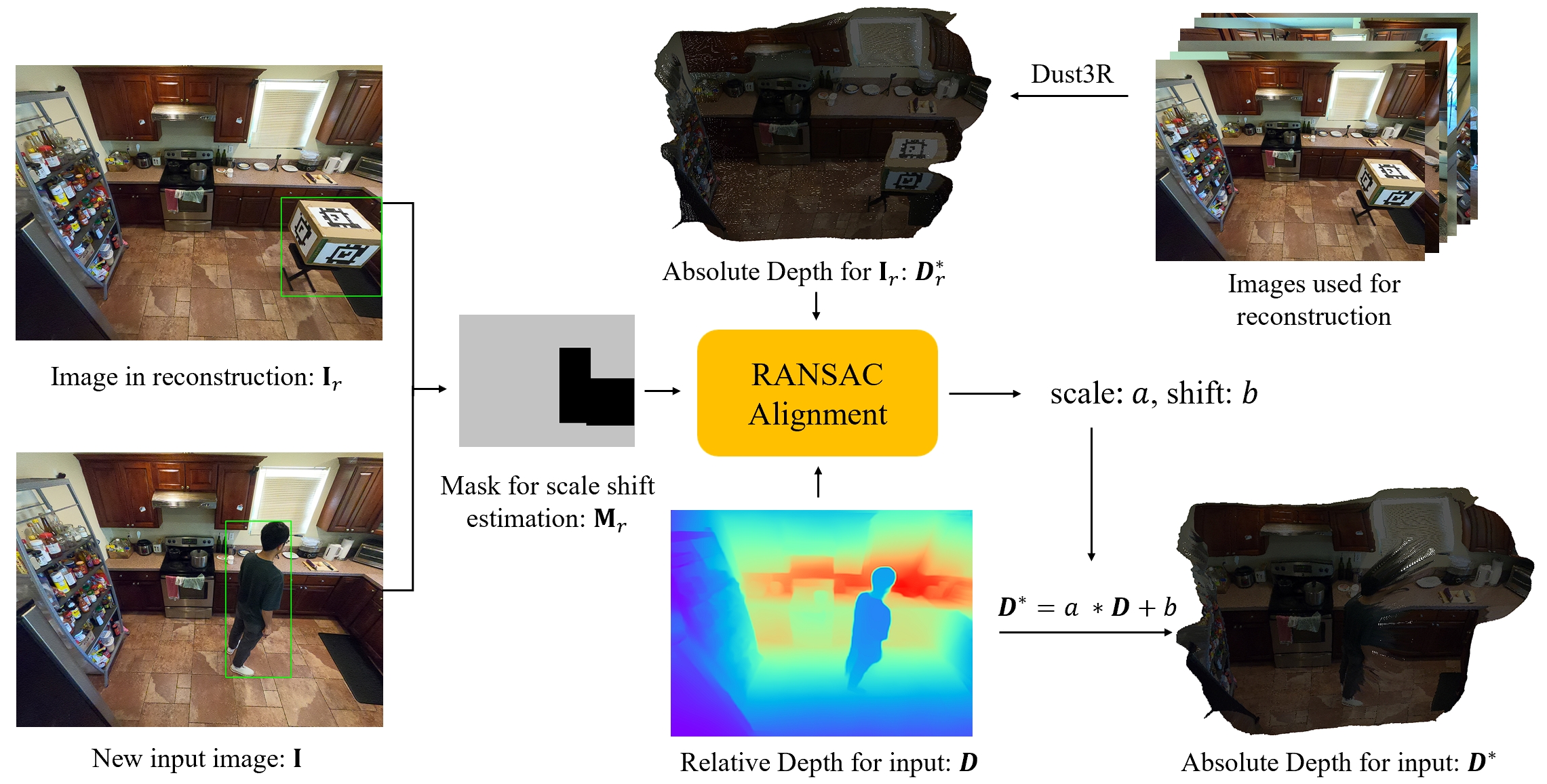}\\
    \caption{}
    \label{fig:absolute_depth_est}
    \end{subfigure}
    \begin{subfigure}[b]{\textwidth}
    \centering
    \includegraphics[width=0.8\linewidth]{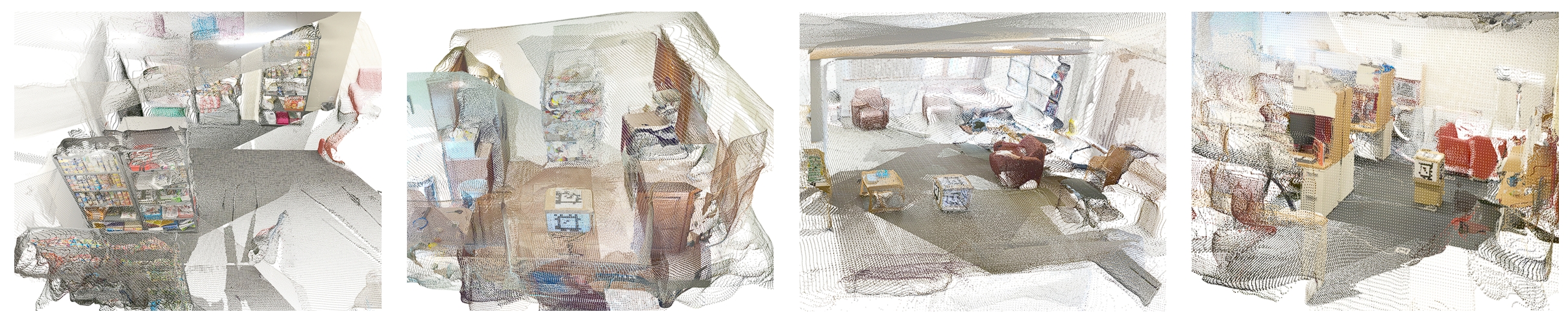}
    \caption{}
    \label{fig:recontruct_results}
    \end{subfigure}
    \centering
    \caption{(a). The procedure of absolute depth estimation. We first reconstruct the scene and obtain the absolute depth for all views using a set of images from 6 cameras (e.g., calibration images). For a new input image, we estimate the scale and shift between the monocular depth map and the absolute depth obtained from reconstruction using RANSAC, by masking out the person bounding box location along with the calibration cube location in the reconstruction image. We use the estimated scale and shift values to obtain the absolute depth for the new input image. (b) Example point clouds of the entire scenes reconstructed by Dust3R \cite{wang2024dust3r} using 6 images from 6 cameras.}
    \vspace{-0.2cm}
    \label{fig:3d_overall}
\end{figure*}

\section{Extension to More Camera Views} \label{sec:view_extend}

\begin{table}[h]
\centering
\begin{tabular}{l c c c}
\Xhline{2\arrayrulewidth}
Method & Views &  Dist. $\downarrow$  & AP $\uparrow$ \\
\Xhline{\arrayrulewidth}
Ours-Single & 1  &  0.150  &  0.868  \\
Ours & 2 & 0.130 & 0.894 \\
Ours & 4 & 0.121 & 0.897 \\
Ours & 6 & \textbf{0.118} & \textbf{0.898}\\
\Xhline{2\arrayrulewidth}
\end{tabular}
\vspace{0.4cm}
\caption{Results of extending to more camera views. Our method shows greater improvements when using more camera views.}
\label{tab:cross_view_results}
\end{table}

In this section, we extend our method to more than 2 camera views with a simple strategy by using multiple view pairs with the same primary view image simultaneously. E.g., when using 4 camera views, we use 3 view pairs where the same primary view (e.g. Camera1) is paired with 3 different reference view images (e.g., Cam2, Cam3, Cam4). For each pair with index $i$, we obtain an uncertainty score $\sigma^{i}$ from the view with the lower $\sigma$ value (i.e., the view selected in UGS). From all 3 pairs with their obtained uncertainty scores $\sigma^{1}, \sigma^{2}, \sigma^{3}$, we choose the output from the pair with the lowest $\sigma$ value as the prediction for the primary view. This enables our method to leverage more camera views without any additional training.

Tab.\ref{tab:cross_view_results} shows the results. Unlike Tab.1 in the main paper, here we investigate the effect of the number of camera views used without considering the head and gaze target visibility of the reference views. In this way, a primary view image is paired with each of the 5 other camera views and forms 5 pairs. In evaluation, for each primary view image, we randomly select 1 pair when using 2 views, 3 pairs when using 4 views, and all 5 pairs when using 6 views. When using 2 or 4 camera views, we run the model 5 times and report the average performance. Although our method with two views already outperforms the single-view baseline significantly, using more views shows even further improvement.

\section{Absolute Depth Computation} \label{sec:absolute_depth}

In this section, we describe our procedure for estimating the absolute depth in cross-view GTE. As shown in Fig.\ref{fig:absolute_depth_est}, we first reconstruct the 3D scene using a set of images from all 6 cameras (e.g. images used in calibration) and obtain the absolute depth $\bm{D}_r^{*}$ for each camera view. As we mentioned in the main paper, by inputting the camera parameters calibrated in real-world metrics, Dust3R can generate depth estimations that are very close to the absolute depth values by optimizing a reconstruction loss. After the 3D reconstruction, when a new input image comes during training and evaluation, we estimate the scale and shift between its relative depth map $\bm{D}$ from a monocular depth estimation model and the absolute depth $\bm{D}_r^{*}$ obtained from reconstruction by masking out the area that changes with a mask $\vect{M}_r$. We use RANSAC to estimate the scale and shift for the input image with this mask $\vect{M}_r$:
 \begin{equation}
     a, b = \underset{\vect{M}_r^{(i,j)}==1}{RANSAC}({\bm{D}}(i,j), \bm{D}_r^*(i,j)),
 \end{equation}

With the estimated scale and shift, we can obtain the absolute depth $\bm{D}^{*}$ of the new input image: $\bm{D}^{*} = a * \bm{D} + b$. Note that after reconstruction, we do not require additional camera views to be available for input. We run this procedure for both the primary and reference views. With the absolute depth from both views obtained, we can obtain the real 3D eye location in the reference view, and transform the location to the primary view's coordinate system using the extrinsic parameters. In this way, the FoV heatmap for the primary view can be obtained in cross-view GTE settings, even when the person is not visible in the primary view.

\section{Camera Calibration Details} \label{sec:calibration}

\begin{figure}[h]
\centering
\setlength\tabcolsep{1.0pt}
\begin{tabular}{ccc}
\includegraphics[width=0.46\linewidth]{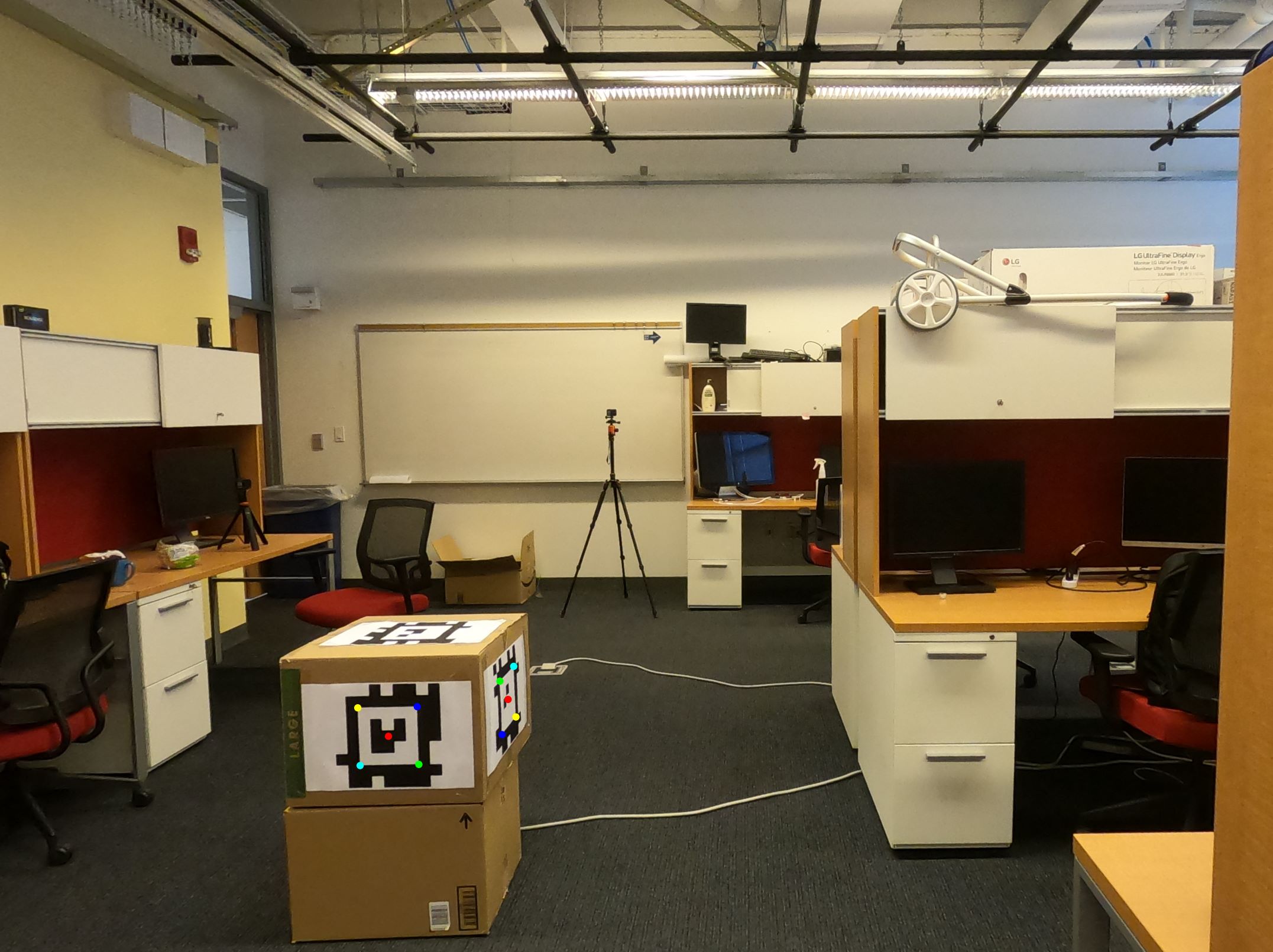}&
 \includegraphics[width=0.46\linewidth]{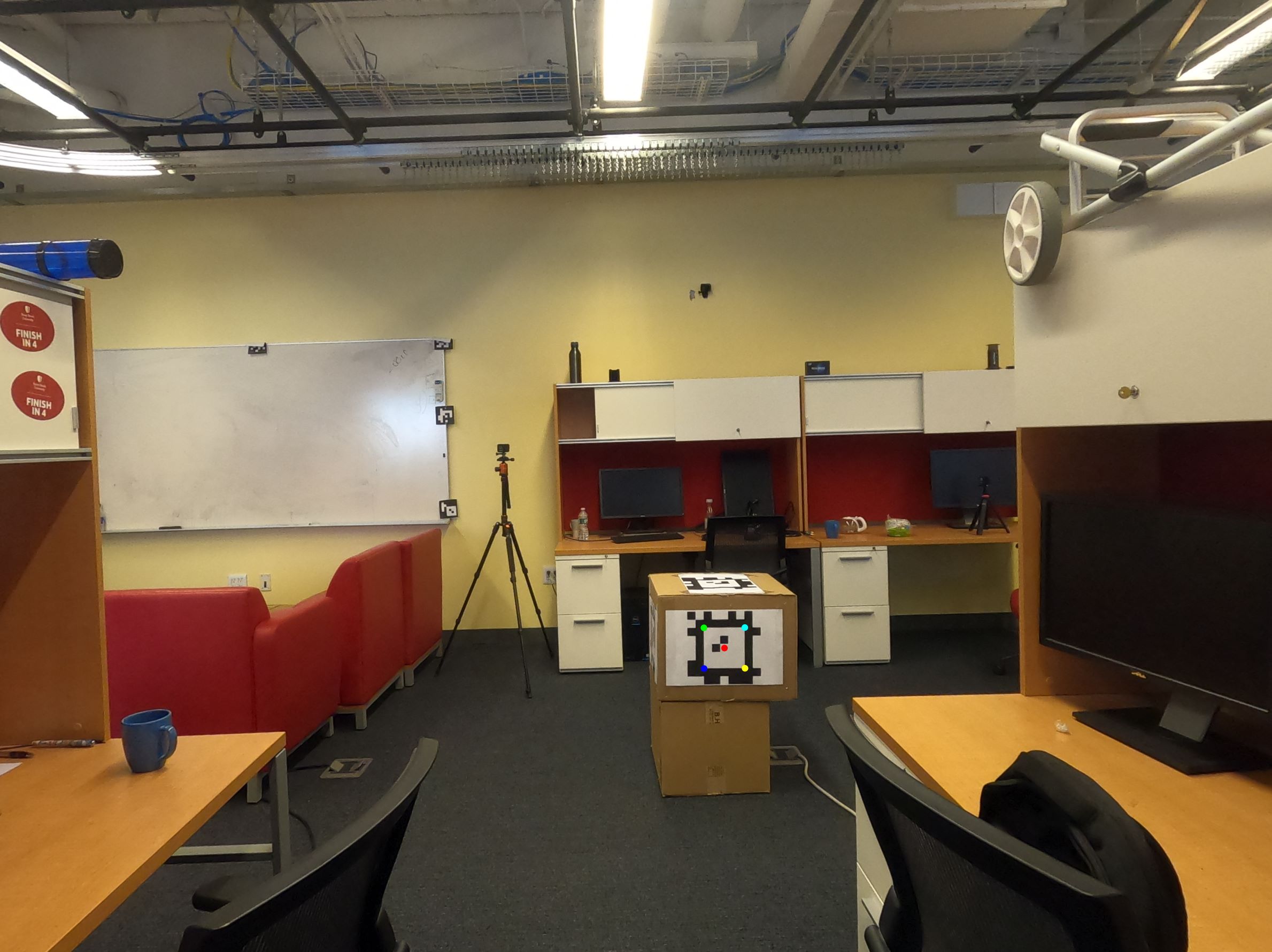}\\
 \includegraphics[width=0.46\linewidth]{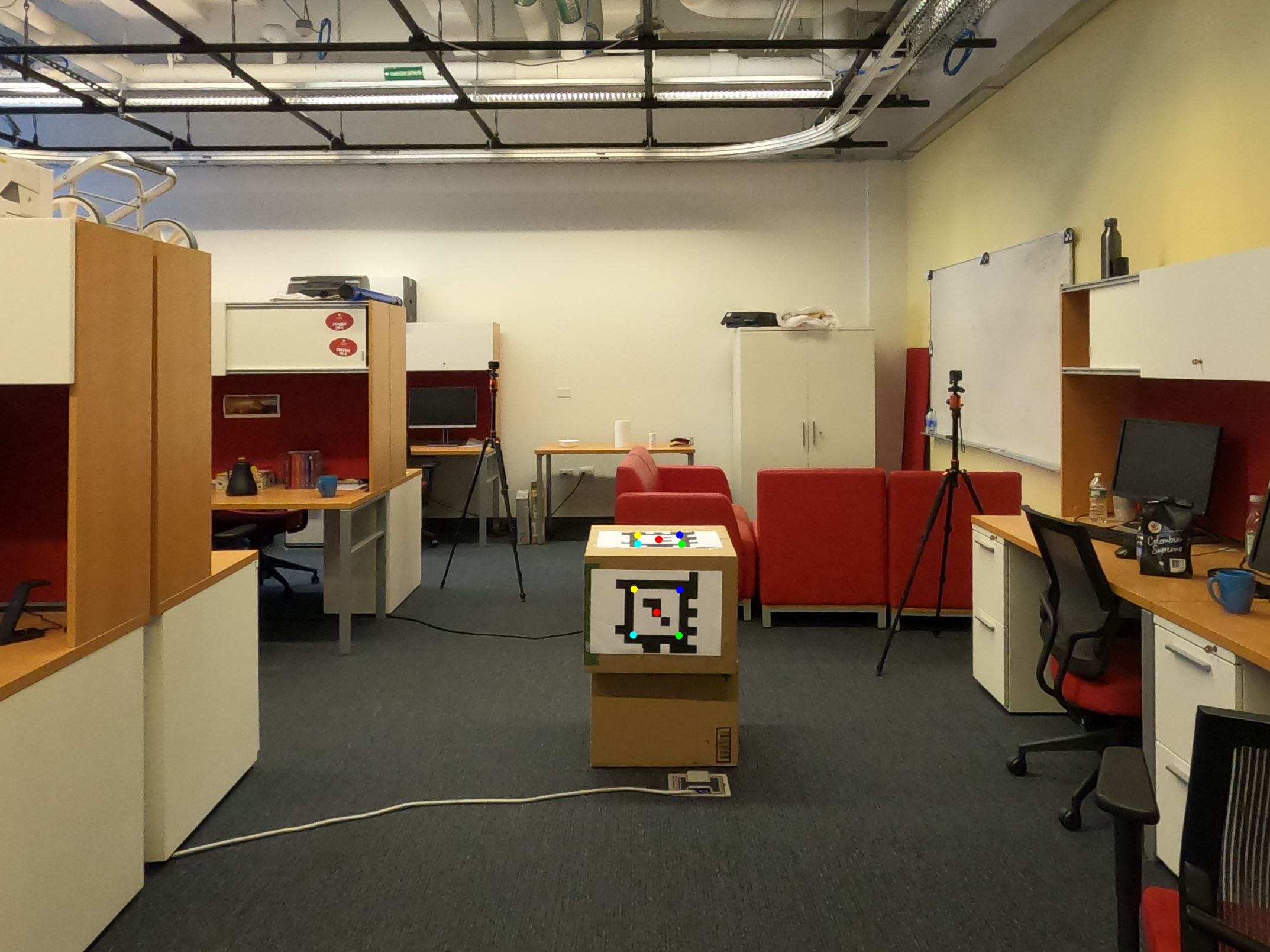}&
\includegraphics[width=0.46\linewidth]{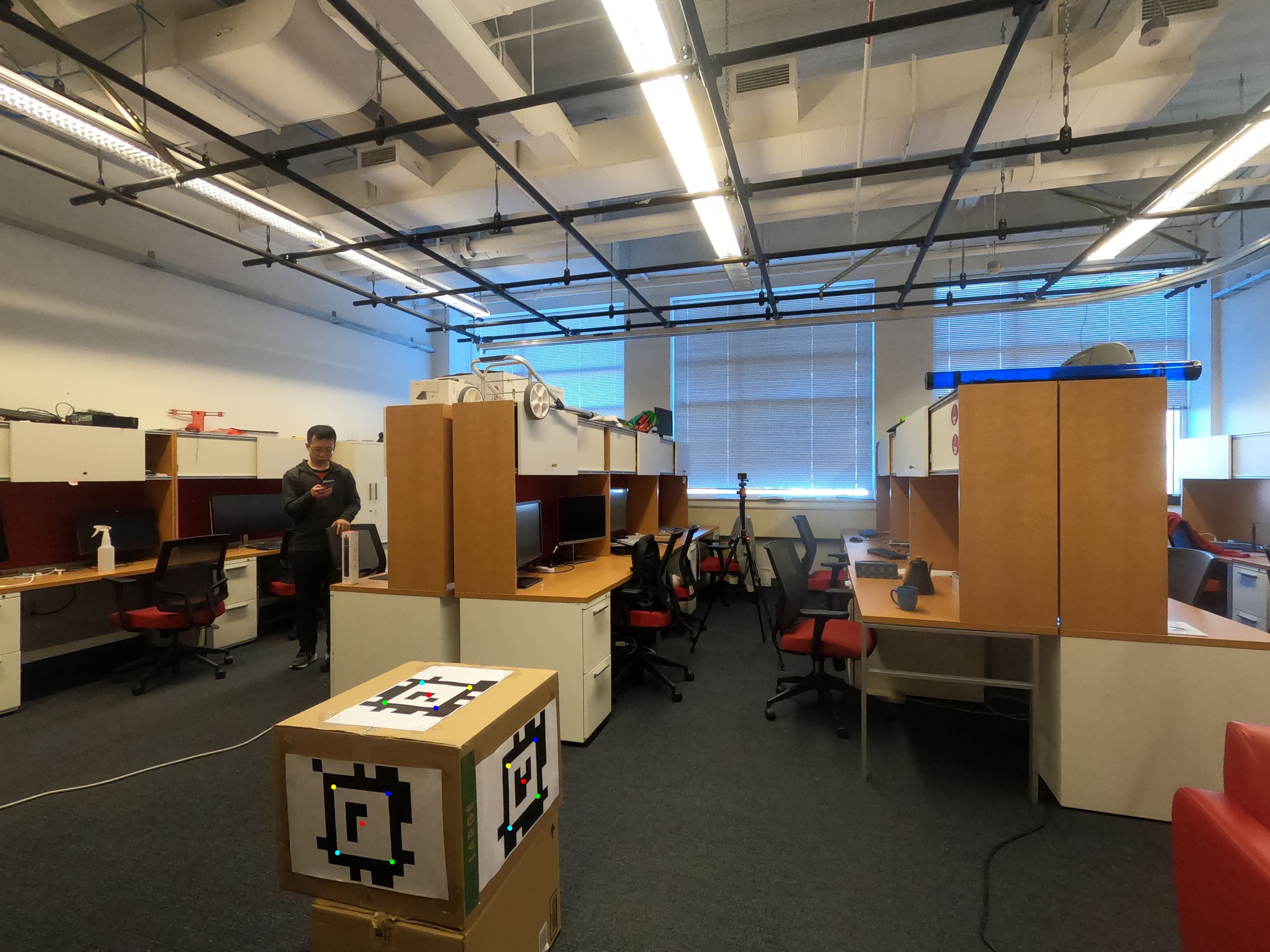}\\
 \includegraphics[width=0.46\linewidth]{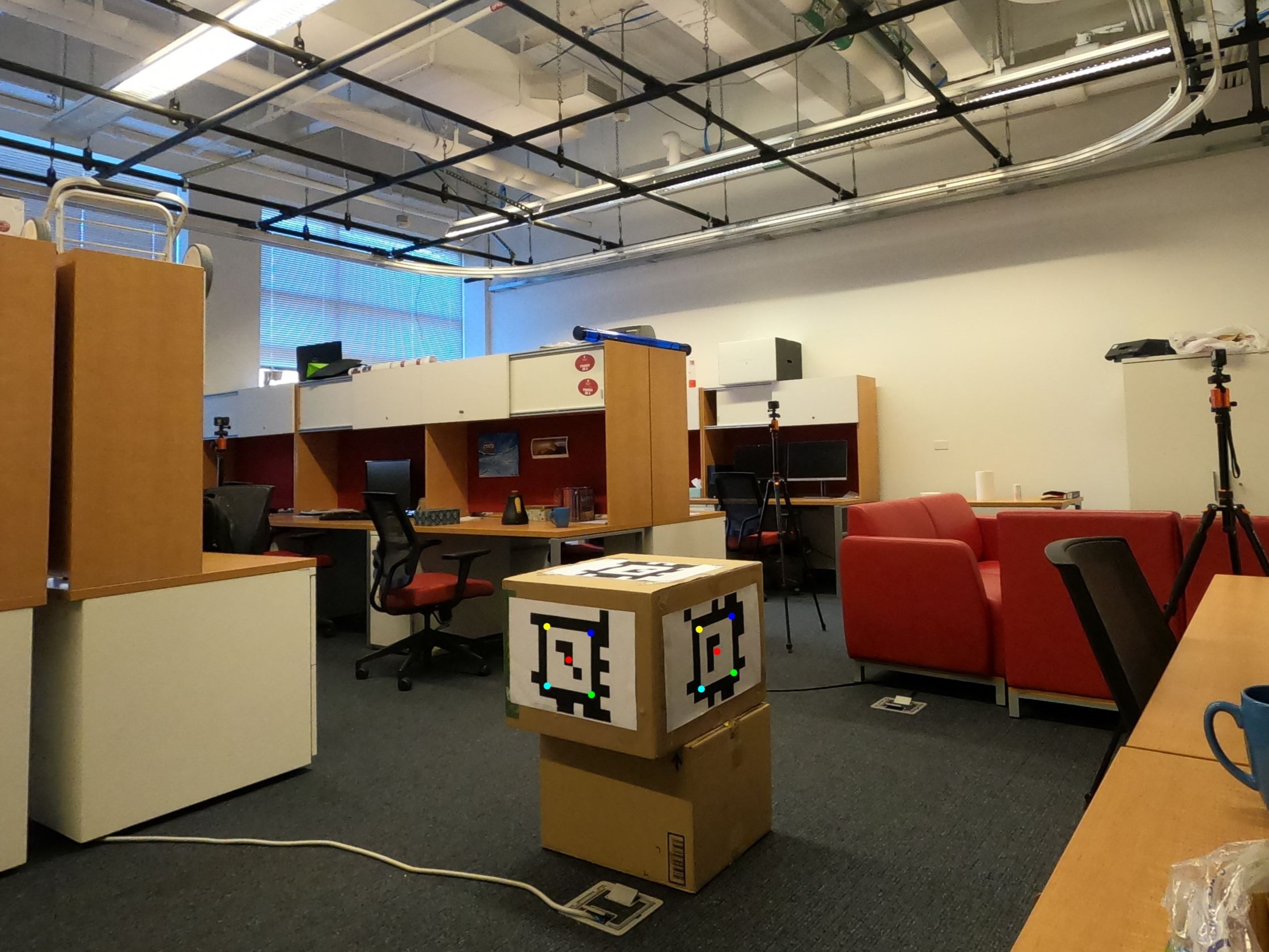}&
 \includegraphics[width=0.46\linewidth]{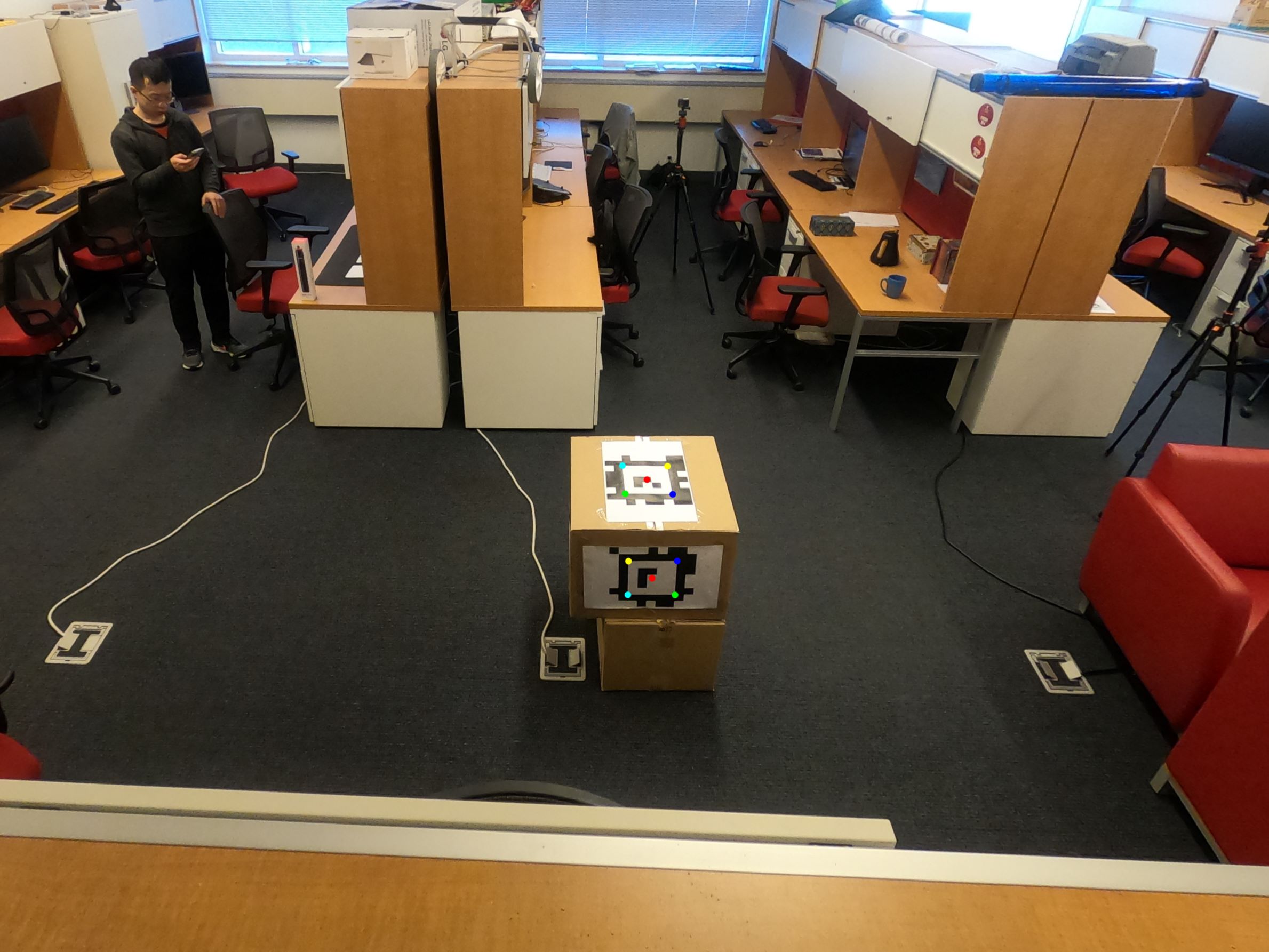}\\

\end{tabular}
\caption{Protocol for calibrating extrinsic camera parameters. We use a cube stuck with AprilTag patterns to calibrate all cameras' extrinsic parameters. The automatically detected corners for the patterns are visualized. We measured the 3D physical locations of these corners before calibration.}
\label{fig:extri_calibration}
\end{figure}

In this section, we provide more details of our camera calibration procedure. We used chessboard patterns for calibrating the intrinsic parameters, which is a common protocol for intrinsic parameter calibration. For calibrating the extrinsic camera parameters, we used a cube stuck with AprilTag patterns on different faces. As shown in Fig.\ref{fig:extri_calibration}, we first measure the 3D physical locations of all the AprilTag corners in the cube, and put the cube in a location where it can be observed by all cameras. The 2D locations of the corners can be automatically detected \cite{olson2011apriltag}, and the extrinsic parameters can be obtained using PnP (Perspective-nPoints). By using this protocol, the extrinsic parameters can be obtained with just one set of images taken. We calibrate the extrinsic parameters before the data collection session of each subject. If no position is visible to all cameras without occlusion, calibration is performed twice using a common camera as the shared coordinate system.

\section{Analyses of Predicted Gaze Vectors} \label{sec:angular_error}

{In this section, we specifically investigate the angular errors of the predicted gaze vectors in our model under different ablation conditions. The ground truth gaze vectors are obtained from the 3D eye locations and gaze target locations computed from triangulation when they appear in multiple camera views. This provides more accurate ground truths than the ``pseudo" gaze vectors used in training.  Tab.\ref{tab:ang_err} shows the results. Training with the uncertainty loss results in an overall improvement in predicted gaze vectors. The HIA leads to a significant improvement when the head is visible, showing its effectiveness in aggregating head information from another view. It also shows a small improvement when the head is not visible. We hypothesize that this is because the model benefits from the overall multi-view training. The UGS module shows improvement when the reference view includes the head but not the gaze target, where the subject is typically facing the camera with a clearly visible face and is selected as the more reliable gaze vector. The ESA module is responsible for aggregating scene information, and does not show a large change in gaze vector prediction.} 

\begin{table}[h]
\centering
\setlength{\tabcolsep}{4pt}
\small
\begin{tabular}{cccc|cc|cc}
\Xhline{2\arrayrulewidth}
\multirow{4}{*}{$\sigma$} & \multirow{4}{*}{HIA} & \multirow{4}{*}{UGS} & \multirow{4}{*}{ESA} & \multicolumn{2}{c|}{Head Vis.}                                                            & \multicolumn{2}{c}{Head Not Vis.}                                                            \\ \cline{5-8}
  &  & & & \multicolumn{1}{c}{\begin{tabular}[c]{@{}c@{}}Target \\ Vis.\end{tabular}} & \begin{tabular}[c|]{@{}c@{}}Target \\ Not Vis.\end{tabular} & \multicolumn{1}{c}{\begin{tabular}[c]{@{}c@{}}Target \\ Vis.\end{tabular}} & \begin{tabular}[c|]{@{}c@{}}Target \\ Not Vis.\end{tabular}       \\ \cline{5-8} 
    &  & & &  \multicolumn{1}{c}{Ang. $\downarrow$} & \multicolumn{1}{c|}{Ang. $\downarrow$} &\multicolumn{1}{c}{Ang. $\downarrow$} & \multicolumn{1}{c}{Ang. $\downarrow$}\\ \Xhline{2\arrayrulewidth}
&  & & &  28.71   & \multicolumn{1}{c|}{28.93}   &   29.15  &  29.73  \\ 
\checkmark &  & & &        26.94                 & \multicolumn{1}{c|}{27.28}   &   27.69                       & 28.22   \\
\checkmark &  \checkmark & & &        21.20                  & \multicolumn{1}{c|}{21.32}   &   25.22                       & 26.40    \\
\checkmark &  \checkmark & \checkmark & &    \underline{20.50}                    & \multicolumn{1}{c|}{\textbf{19.72}}  & \underline{24.87} & \textbf{25.78}  \\
\checkmark&  \checkmark & \checkmark & \checkmark &       \textbf{20.47}                   & \multicolumn{1}{c|}{\underline{20.02}}   &         \textbf{24.68}                 &   \underline{25.84} \\ \Xhline{2\arrayrulewidth}
\end{tabular}
\caption{Angular Errors of predicted gaze vectors of our method under different ablation conditions.}
\label{tab:ang_err}
\end{table}

\section{Analyses of HIA Module} \label{sec:hia_analysis}

In this section, we provide more detailed analyses of the HIA module. We first analyzed the effects of the two additional inputs of HIA: the head crop image in the other view, and the relative rotations between views computed from the extrinsic parameters. {Tab.\ref{tab:hia_ablation} shows the results. Here we treat our method only with the HIA module (without UGS and ESA) as the base model to avoid the potential influence of other components. In the 2nd row, we discard the head image input in the reference view, by inputting zero-padded tensors to the HIA module in all cases. It shows a large drop in all metrics when the head is visible, demonstrating that HIA effectively leverages the head appearances from the other view to enhance the embeddings. The performance does not change when the head is not visible, as the input is the same. In the first row, we trained the model with HIA by removing the input camera parameters (relative rotations). This also results in a significant drop when the head is visible. This suggests that without the geometric relationship between views, the model cannot learn how to effectively uses the head appearance from the other view. There is also a large drop in AP when the head is not visible in the reference view. This shows that with the input camera parameters, the head embedding is enhanced with 3D geometric-aware information and benefits the performance of in/out prediction when input to the in/out prediction head, as shown in Fig.4 in the main paper. }

\begin{table}[h]
\begin{floatrow}
\centering
\footnotesize
\setlength\tabcolsep{2.5pt}
\begin{tabular}{l|cccc|cccc}
\Xhline{2\arrayrulewidth}
\multirow{3}{*}{Method} & \multicolumn{4}{c|}{Head Vis.}                                                            & \multicolumn{4}{c}{Head Not Vis.}                                                            \\ 
  & \multicolumn{2}{c}{Target Vis.}                            & \multicolumn{2}{c|}{Target Not Vis.}      & \multicolumn{2}{c}{Target Vis.}                            & \multicolumn{2}{c}{Target Not Vis.}       \\ \cline{2-9} 
     & \multicolumn{1}{c}{Dist. $\downarrow$} & \multicolumn{1}{c}{AP $\uparrow$} & \multicolumn{1}{c}{Dist. $\downarrow$} & AP $\uparrow$ & \multicolumn{1}{c}{Dist. $\downarrow$} & \multicolumn{1}{c}{AP $\uparrow$} & \multicolumn{1}{c}{Dist. $\downarrow$} & AP $\uparrow$ \\ \Xhline{2\arrayrulewidth}
No Cam. &   0.150            &  \multicolumn{1}{c|}{0.880}     &     0.152                   &  0.875  &    0.175     & \multicolumn{1}{c|}{0.776}      &         \textbf{0.154}                &  0.847 \\
No Head &   0.149                     & \multicolumn{1}{c|}{0.876}   &     0.155                   &  0.886  &    \textbf{0.174}        & \multicolumn{1}{c|}{\textbf{0.821}}          &        0.155                &  \textbf{0.873} \\
Ours-HIA     &       \textbf{0.135}     & \multicolumn{1}{c|}{\textbf{0.896}}        &         \textbf{0.133}                 &   \textbf{0.897}   & \textbf{0.174}    &  \multicolumn{1}{c|}{\textbf{0.821}}       &        0.155                &  \textbf{0.873}   \\ \Xhline{2\arrayrulewidth}
\end{tabular}
{\caption{Analysis of the HIA module. We experimented by discarding the head crop input or training without inputting camera parameters, using our model with only HIA as the base model.} \label{tab:hia_ablation}}
\end{floatrow}
\end{table}

We also explored alternative strategies for aggregating head features and training the gaze estimator. In Tab. \ref{tab:multiply_and_fix} (Row 1), we replaced concatenation with multiplication when including the camera rotation matrix in HIA's cross-attention, which led to worse performance, especially when the head is not visible in the reference view. In Row 2, fixing the gaze backbone during training also resulted in significantly degraded performance.

\begin{table}[h]
\centering
\footnotesize
\setlength\tabcolsep{2.5pt}
\begin{tabular}{l|cccc|cccc}
\Xhline{2\arrayrulewidth}
\multirow{3}{*}{Method} & \multicolumn{4}{c|}{Head Vis.}                                                            & \multicolumn{4}{c}{Head Not Vis.}                                                            \\ 
  & \multicolumn{2}{c}{Target Vis.}                            & \multicolumn{2}{c|}{Target Not Vis.}      & \multicolumn{2}{c}{Target Vis.}                            & \multicolumn{2}{c}{Target Not Vis.}       \\ \cline{2-9} 
     & \multicolumn{1}{c}{Dist. $\downarrow$} & \multicolumn{1}{c|}{AP $\uparrow$} & \multicolumn{1}{c}{Dist. $\downarrow$} & AP $\uparrow$ & \multicolumn{1}{c}{Dist. $\downarrow$} & \multicolumn{1}{c|}{AP $\uparrow$} & \multicolumn{1}{c}{Dist. $\downarrow$} & AP $\uparrow$ \\ \Xhline{2\arrayrulewidth}
Ours-Mul &     0.131                    & \multicolumn{1}{c|}{0.903}   &          0.124               &  0.904  &    0.172                 & \multicolumn{1}{c|}{0.801}   &        0.160                & 0.860  \\

Ours-Fix &     0.145                    & \multicolumn{1}{c|}{0.894}   &          0.139               &  0.897  &    0.193                 & \multicolumn{1}{c|}{0.805}   &        0.171                & 0.841  \\
Ours     &       \textbf{0.129}                   & \multicolumn{1}{l|}{\textbf{0.909}}   &         \textbf{0.122}                 &   \textbf{0.912}  &        \textbf{0.161}                  & \multicolumn{1}{c|}{\textbf{0.836}}   &         \textbf{0.152}                 &  \textbf{0.868}   \\ \Xhline{2\arrayrulewidth}
\end{tabular}
\caption{Results of using multiplication for the relative rotation (Row 1) and fixing the gaze backbone during training (Row 2).}
\label{tab:multiply_and_fix}
\end{table}

\section{Analyses of UGS Module} \label{sec:ugs_analysis}

In this section, we analyze the UGS module in detail by demonstrating the correlation between the error in gaze vector prediction and the predicted uncertainty score $\sigma$, the performance of UGS on samples with large errors in predicted gaze vectors,  and showing the effect of the UGS module with some qualitative examples.

\begin{figure}[h]
    \centering
    \includegraphics[width=0.6\linewidth]{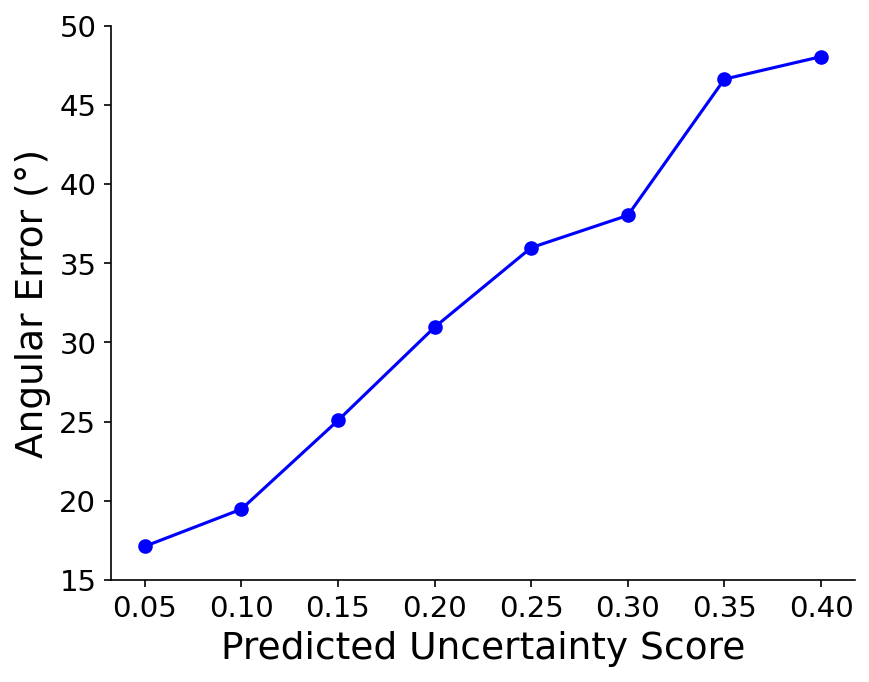}
    \caption{Average angular error for the predicted gaze vectors with their uncertainty scores $\sigma$ falling into different slots. We divide the slot of $\sigma$ by 0.05. The gaze vectors with larger predicted uncertainty scores tend to have larger angular errors.}
    \label{fig:err_with_var}
\end{figure}

In Fig.\ref{fig:err_with_var} we visualize the average angular error of the predicted gaze vectors with their predicted uncertainty scores $\sigma$ falling into different slots. About 93\% of all samples have a predicted $\sigma < 0.2$. It can be seen a larger $\sigma$ value corresponds to a larger error for the gaze vector prediction, supporting our motivation to select the view with a lower $\sigma$ value in an input view pair in the UGS module. 

{We also demonstrated the effectiveness of the UGS module by operating the module on samples with large errors in predicted gaze vectors, of which the predicted gaze vectors have an angular error $>$ 30° before uncertainty-based selection. To show the effectiveness of UGS, the models were evaluated on all view pairs where the reference view contains the head of the subject. As shown Tab.\ref{tab:ugs_largeerror}, in addition to the substantial reduction in angular errors for gaze vectors after selection, the Dist. and Ap. metrics in GTE also exhibit notable improvements, highlighting its crucial impact on samples with large initial prediction errors.}

\begin{table}[h]
\centering
\begin{tabular}{l c c c}
\Xhline{2\arrayrulewidth}
Method & Dist. $\downarrow$  & AP $\uparrow$ & Ang. $\downarrow$ \\
\Xhline{\arrayrulewidth}
No selection & 0.228    & 0.856  &  44.35° \\
Ours  & \textbf{0.200}  & \textbf{0.883} &  \textbf{35.91°} \\
\Xhline{2\arrayrulewidth}
\end{tabular}
\caption{Effect of the UGS module on samples with large errors in predicted gaze vectors in the primary view before selection.}
\label{tab:ugs_largeerror}
\end{table}

\begin{figure*}[t]
    \centering
    \includegraphics[width=0.7\linewidth]{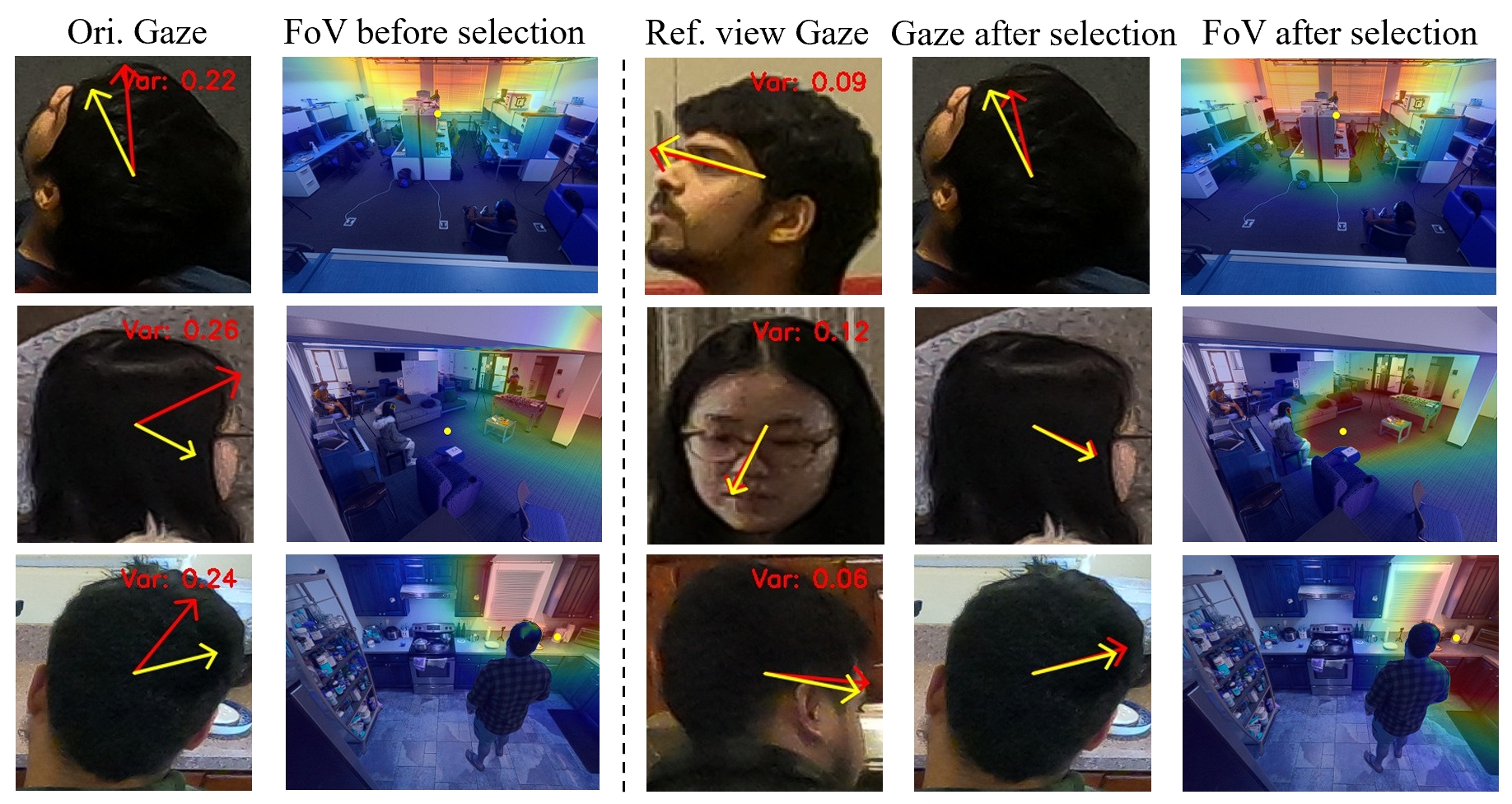}
    \caption{Qualitative examples for the UGS module output. The left side shows the gaze vectors and FoV heatmaps without using UGS, and the right shows the gaze vector and FoV heatmaps after selection and replacement in UGS.  Red vectors correspond to the predicted gaze vectors while yellow ones are the ground truth. Leveraging the reference view with a more accurate gaze vector predicted and a lower uncertainty score, the UGS module can output a FoV heatmap with much better quality.}
    \label{fig:ugs_examples}
\end{figure*}

Fig.\ref{fig:ugs_examples} demonstrates the effect of the UGS module. By leveraging the reference view with a lower $\sigma$ value, the unreliable gaze vector from the primary view will be replaced with a much more accurate gaze vector, and lead to an FoV heatmap with a much better quality.

\section{Analyses of ESA module} \label{sec:esa_analysis}
\begin{figure}[h]
\centering
\setlength\tabcolsep{8.0pt}
\begin{tabular}{cc}
 Query Point &
 Attention Weights\\
\includegraphics[width=0.40\linewidth]{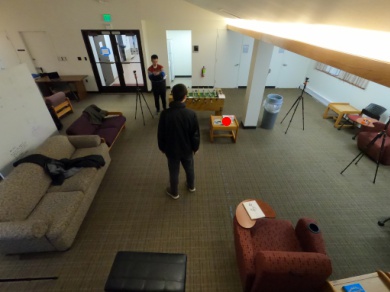}&
 \includegraphics[width=0.40\linewidth]{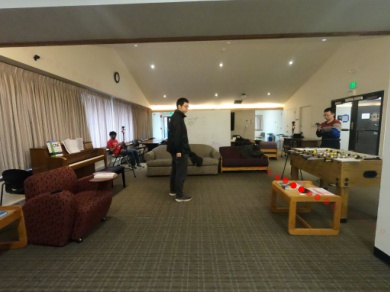} \\
 \includegraphics[width=0.40\linewidth]{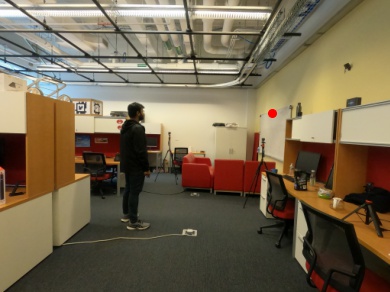}&
 \includegraphics[width=0.40\linewidth]{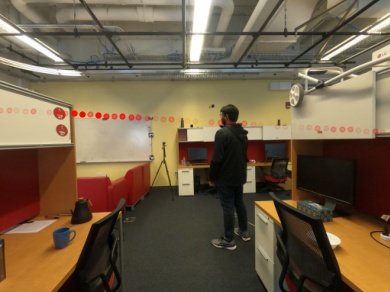} \\
 \includegraphics[width=0.40\linewidth]{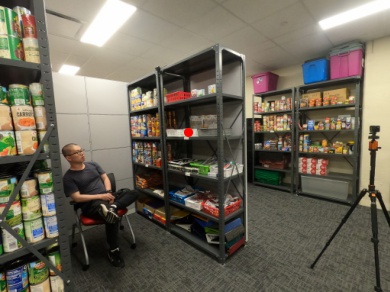}&
 \includegraphics[width=0.40\linewidth]{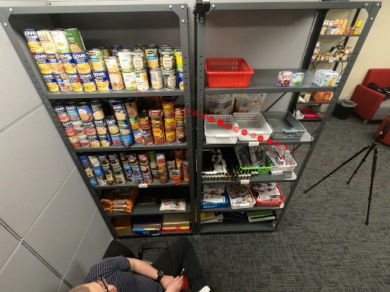} \\
\end{tabular}
\caption{Visualizations of attention weights in the ESA module. The right column visualizes the attention weights between the query point in the left column and the sampled feature tokens along the epipolar line corresponding to the query in the reference view. Larger attention values are shown with higher intensity. The tokens located near the same location as the query point show the highest weights along the entire epipolar line.}
\label{fig:epipolar_weights}
\end{figure}

In this section, we provide more analyses of the ESA module, by showcasing its performance on the samples with occlusion, and visualizing the attention weights of the epipolar attention in the ESA module. We show the effectiveness of the ESA module by visualizing the epipolar attention weights. As explained in the main paper, each feature token in one view will be engaged in cross attention with feature tokens sampled along the epipolar line in the other view. In Fig.\ref{fig:epipolar_weights}, we select a query feature location in the primary view and visualize the attention weights with the sampled feature tokens along the epipolar line in the reference view. We average the attention weights across all heads in the cross-attention module. It can be seen the attention weights are the highest for tokens located near the same location as the query point, demonstrating the effectiveness of the ESA module in aggregating useful scene background information from the other view.

{On the other hand, we also perform an ablation study of the ESA module on samples which are occluded in the primary view, to investigate the effectiveness of epipolar attention in differentiating occluding objects by using information from another view. We manually annotated the locations of the occluded samples in the primary view by referring to the target locations (laser points) in other views, making up 1576 pairs of input views. Most of these targets are partially occluded by another object, or self-occluded, i.e., the laser point is located on the invisible side of the object. As the total number of occluded samples are relatively small and we assigned all occlusion samples to the ``inside" class for the in/out task, we observed the model showing performance close to 1 in AP. Therefore, we just evaluated the Dist. metric for the GTE task.}

{Tab.\ref{tab:esa_occlusion} shows the results. Compared to the model without ESA module, the full model shows an obvious performance when the target is visible in the reference view. This supports our claim in the main paper that ESA provides complementary information on the potential gaze object from the reference view, especially in disambiguating the gaze target in case of occlusion in the primary view.}

\begin{table}[h]
\begin{floatrow}
\centering
\footnotesize
\setlength\tabcolsep{4.0pt}
\begin{tabular}{l|cc|cc}
\Xhline{2\arrayrulewidth}
\multirow{3}{*}{Method} & \multicolumn{2}{c|}{Head Vis.}                                                            & \multicolumn{2}{c}{Head Not Vis.}                                                            \\ 
  & \multicolumn{1}{c}{Target Vis.}                            & \multicolumn{1}{c|}{Target Not Vis.}      & \multicolumn{1}{c}{Target Vis.}                            & \multicolumn{1}{c}{Target Not Vis.}       \\ \cline{2-5} 
     & \multicolumn{1}{c}{Dist. $\downarrow$} & \multicolumn{1}{c|}{Dist. $\downarrow$}  & \multicolumn{1}{c}{Dist. $\downarrow$} & \multicolumn{1}{c}{Dist. $\downarrow$}  \\ \Xhline{2\arrayrulewidth}
No ESA &       0.161           &  \textbf{0.152}              &      0.171               &   0.131              \\
Ours     &       \textbf{0.150}                     &         0.155                 &    \textbf{0.162}                     &         \textbf{0.130}                \\ \Xhline{2\arrayrulewidth}
\end{tabular}
{\caption{Ablation of ESA module on samples with occlusion. Without ESA, the model shows obvious drop in performance when the target is visible in the reference view.} \label{tab:esa_occlusion}}
\end{floatrow}
\end{table}


\section{Sensitivity Analysis to Camera Parameters} \label{sec:camera_sensitivity}

In the main paper, we provide the results of our model with calibrated camera parameters. In this section, we analyze the sensitivities of our model to changes in camera parameters by considering potential errors and noise in camera calibration in real applications. We randomly jittered the intrinsic and extrinsic parameters by -5\% $\sim$ 5\%, and show the results of the multi-view and cross-view experiments in Tab.\ref{tab:camparam_sensitivity} and Tab.\ref{tab:camparam_sensitivity_cross}. As shown in both tables, the model only has a slight drop in performance with the jittered camera parameters in both general multi-view and cross-view tasks. 

\begin{table}[h]
\begin{floatrow}
\centering
\footnotesize
\setlength\tabcolsep{2.5pt}
\begin{tabular}{l|cccc|cccc}
\Xhline{2\arrayrulewidth}
\multirow{3}{*}{Method} & \multicolumn{4}{c|}{Head Vis.}                                                            & \multicolumn{4}{c}{Head Not Vis.}                                                            \\ 
  & \multicolumn{2}{c}{Target Vis.}                            & \multicolumn{2}{c|}{Target Not Vis.}      & \multicolumn{2}{c}{Target Vis.}                            & \multicolumn{2}{c}{Target Not Vis.}       \\ \cline{2-9} 
     & \multicolumn{1}{c}{Dist. $\downarrow$} & \multicolumn{1}{c|}{AP $\uparrow$} & \multicolumn{1}{c}{Dist. $\downarrow$} & AP $\uparrow$ & \multicolumn{1}{c}{Dist. $\downarrow$} & \multicolumn{1}{c|}{AP $\uparrow$} & \multicolumn{1}{c}{Dist. $\downarrow$} & AP $\uparrow$ \\ \Xhline{2\arrayrulewidth}
Ours* &   0.130                      & \multicolumn{1}{c|}{0.907}   &      0.124                    &  0.910  &     0.165                   & \multicolumn{1}{c|}{0.830}   &         0.153                &  0.865  \\
Ours     &       \textbf{0.129}                   & \multicolumn{1}{l|}{\textbf{0.909}}   &         \textbf{0.122}                 &   \textbf{0.912}  &        \textbf{0.161}                  & \multicolumn{1}{c|}{\textbf{0.836}}   &         \textbf{0.152}                 &  \textbf{0.868}   \\ \Xhline{2\arrayrulewidth}
\end{tabular}
{\caption{Results of sensitivity analyses to camera parameters. Ours* shows the performance of the model with camera parameters jittered by -5\% $\sim$ 5\%, and Ours is the original model. Our model shows little drop in performance in all conditions.} \label{tab:camparam_sensitivity}}
\end{floatrow}
\end{table}

\begin{table}[h]
\centering
\small
\begin{tabular}{l c c}
\Xhline{2\arrayrulewidth}
Method & Dist. $\downarrow$  & AP $\uparrow$ \\
\Xhline{\arrayrulewidth}
Ours* & 0.190  & 0.817 \\
Ours  & \textbf{0.188}  & \textbf{0.820} \\
\Xhline{2\arrayrulewidth}
\vspace{-8pt}
\caption{Results of sensitivity analyses to camera parameters for the cross-view task.}
\label{tab:camparam_sensitivity_cross}
\end{tabular}
\end{table}

\section{Model Complexity and Cost} \label{sec:modelcost}
Tables~\ref{tab:multi_view_cost} and \ref{tab:cross_view_cost} present the model size and inference speed on an RTX A5000 GPU for the multi-view and cross-view settings. Our model has a relatively larger number of parameters compared to single view baselines due to the use of transformer \cite{vit, vaswani2017attention, zhao2024itransformer} as the encoder. However, our method only adds a very small number of parameters for multi-view processing compared to Ours-single, and it shows a large improvement in performance. Regarding inference speed, our method runs at 61.53 ms per image (16.25 FPS), which is acceptable considering the theoretical lower bound is 2 x due to processing two images. Despite inference speed not being a primary design goal, we believe real-time performance is achievable with techniques like quantization.

\begin{table}[h]
\captionsetup{font=footnotesize}
\begin{floatrow}
\centering
\setlength{\tabcolsep}{0.45pt}
\capbtabbox[0.45\textwidth]{
\scriptsize
\begin{tabular}{l c c c}
\Xhline{2\arrayrulewidth}
\multirow{2}{*}{Methods} & \multirow{2}{*}{Params.} & Runtime & Dist. $\downarrow$  \\
       &         &      (ms)        & (Head Visible) \\
\Xhline{\arrayrulewidth}
Chong\cite{chong2020detecting} & 61.5M  & 14.92 & 0.158 \\
Miao\cite{miao2023patch} & 61.7M   & 15.25  & 0.141\\
Tafasca*\cite{tafasca2023childplay} & 21.7M  & 16.74 & 0.148\\
Ours-single & 101.9M & 20.63 & 0.150\\
Ours  & 107.6M & 61.53& \textbf{0.125} \\
\Xhline{2\arrayrulewidth}
\end{tabular}
}
{
\caption{\# Parameters and runtime of multi-view models. }
\label{tab:multi_view_cost}
}
\capbtabbox[0.45\textwidth]{
\scriptsize
\begin{tabular}{l c c c c}
\Xhline{2\arrayrulewidth}
\multirow{2}{*}{Methods} & \multirow{2}{*}{Params.} & Runtime & \multirow{2}{*}{Dist$\downarrow$} \\
 &         &      (ms)        & \\
\Xhline{\arrayrulewidth}
DeepGazeIIE\cite{deepgazeiie} & 104.1M  & 248.76 & 0.248\\
Recasens\cite{recasens2017following} & 189.3M & 10.48 & 0.271\\
Ours  & 108.4M & 62.21 & \textbf{0.188}\\
\Xhline{2\arrayrulewidth}
\end{tabular}
}
{
\caption{\# Parameters and runtime of cross-view models}
\label{tab:cross_view_cost}
}
\end{floatrow}
\end{table}

\section{Training with the same learning rate} \label{sec:samelr}
{In the experiments, we trained the model with different learning rates when evaluating on different scenes in leave-one-scene-out cross validation. We show the results when training the model with the same learning rate in the general multi-view (Tab.\ref{tab:samelr}) and cross-view (Tab.\ref{tab:samelr_cross}) tasks with a learning rate of $2.5 \times 10^{-6}$ and $1 \times 10^{-7}$ respectively. Even when training with the same learning rate without tuning for each scene specifically, the model only shows a small drop in performance in all conditions.}

\begin{table}[h]
\begin{floatrow}
\centering
\footnotesize
\setlength\tabcolsep{2.5pt}
\begin{tabular}{l|cccc|cccc}
\Xhline{2\arrayrulewidth}
\multirow{3}{*}{Method} & \multicolumn{4}{c|}{Head Vis.}                                                            & \multicolumn{4}{c}{Head Not Vis.}                                                            \\ 
  & \multicolumn{2}{c}{Target Vis.}                            & \multicolumn{2}{c|}{Target Not Vis.}      & \multicolumn{2}{c}{Target Vis.}                            & \multicolumn{2}{c}{Target Not Vis.}       \\ \cline{2-9} 
     & \multicolumn{1}{c}{Dist. $\downarrow$} & \multicolumn{1}{c|}{AP $\uparrow$} & \multicolumn{1}{c}{Dist. $\downarrow$} & AP $\uparrow$ & \multicolumn{1}{c}{Dist. $\downarrow$} & \multicolumn{1}{c|}{AP $\uparrow$} & \multicolumn{1}{c}{Dist. $\downarrow$} & AP $\uparrow$ \\ \Xhline{2\arrayrulewidth}
Ours$\dagger$ &   0.134                     & \multicolumn{1}{c|}{0.905}   &      0.125                    &  0.910  &     0.167                   & \multicolumn{1}{c|}{0.823}   &         0.159                &  0.860  \\
Ours     &       \textbf{0.129}                   & \multicolumn{1}{l|}{\textbf{0.909}}   &         \textbf{0.122}                 &   \textbf{0.912}  &        \textbf{0.161}                  & \multicolumn{1}{c|}{\textbf{0.836}}   &         \textbf{0.152}                 &  \textbf{0.868}   \\ \Xhline{2\arrayrulewidth}
\end{tabular}
{\caption{Results of training with the same learning rate in the general multi-view task. Ours$\dagger$ shows the results of training the models with the same learning rate.} \label{tab:samelr}}
\end{floatrow}
\end{table}
\vspace{-6pt}

\begin{table}[h]
\centering
\small
\begin{tabular}{l c c}
\Xhline{2\arrayrulewidth}
Method & Dist. $\downarrow$  & AP $\uparrow$ \\
\Xhline{\arrayrulewidth}
Ours$\dagger$ & 0.199  & 0.814 \\
Ours  & \textbf{0.188}  & \textbf{0.820} \\
\Xhline{2\arrayrulewidth}
\vspace{-8pt}
\caption{Results of training with the same learning rate in the cross-view task.}
\label{tab:samelr_cross}
\end{tabular}
\end{table}

\section{Reproduced results of Tafasca et. al.} \label{sec:reproduce}
In Tab.\ref{tab:reimp_results}, we show the results in the paper and our reimplemented numbers for Tafasca et.al \cite{tafasca2023childplay} on the GazeFollow dataset \cite{NIPS2015_ec895663}. We can achieve almost the same performance. In Tab.1 in the main paper, the model trained on GazeFollow is fine-tuned on our MVGT dataset for evaluation.

\begin{table}[h]
\centering
\begin{tabular}{l c c c}
\Xhline{2\arrayrulewidth}
Method &  AUC $\uparrow$ & Avg. Dist. $\downarrow$  & Min. Dist $\downarrow$ \\
\Xhline{\arrayrulewidth}
Tafasca \cite{tafasca2023childplay} & 0.936 & 0.125 & 0.064 \\
Tafasca* & 0.935 & 0.124 & 0.063\\
\Xhline{2\arrayrulewidth}
\end{tabular}
\vspace{0.4cm}
\caption{Results of the numbers in the paper and the reproduced results on the GazeFollow dataset for Tafasca et.al \cite{tafasca2023childplay}. $^*$ indicates results for the reimplemented model.}
\label{tab:reimp_results}
\end{table}

\section{Discussions and Limitations} \label{sec:discussions}
In this section, we discuss the applicability and generalization of our dataset and method. Although the MVGT dataset is not as large scale as most available GTE datasets, it is the first GTE dataset that includes valuable multi-view scene/subject information, with calibrated camera parameters and precise gaze target annotations. Furthermore, we also introduce a well-defined, non-intrusive data collection protocol for gathering GTE data with accurate annotations, which is easily applicable to new scenes with multi-view setups, and allows for potential scaling up of the dataset. We hope our dataset will inspire broader 
community contributions to multi-view GTE.

For handling different scenarios in multi-view GTE, we proposed two networks in the general multi-view and cross-view task scenarios. The first model is applicable without any assumptions; the second is modified from the first for cross-view GTE, but can only be used when 3D scene reconstruction is available before GTE. In real applications, when a 3D reconstruction exists, both models can be used: the first model is used when the head is in the primary view, and the second otherwise. If not, users can still use the first, which outperforms single-view methods significantly.

\begin{figure}[h]
\centering
\includegraphics[width=\linewidth]{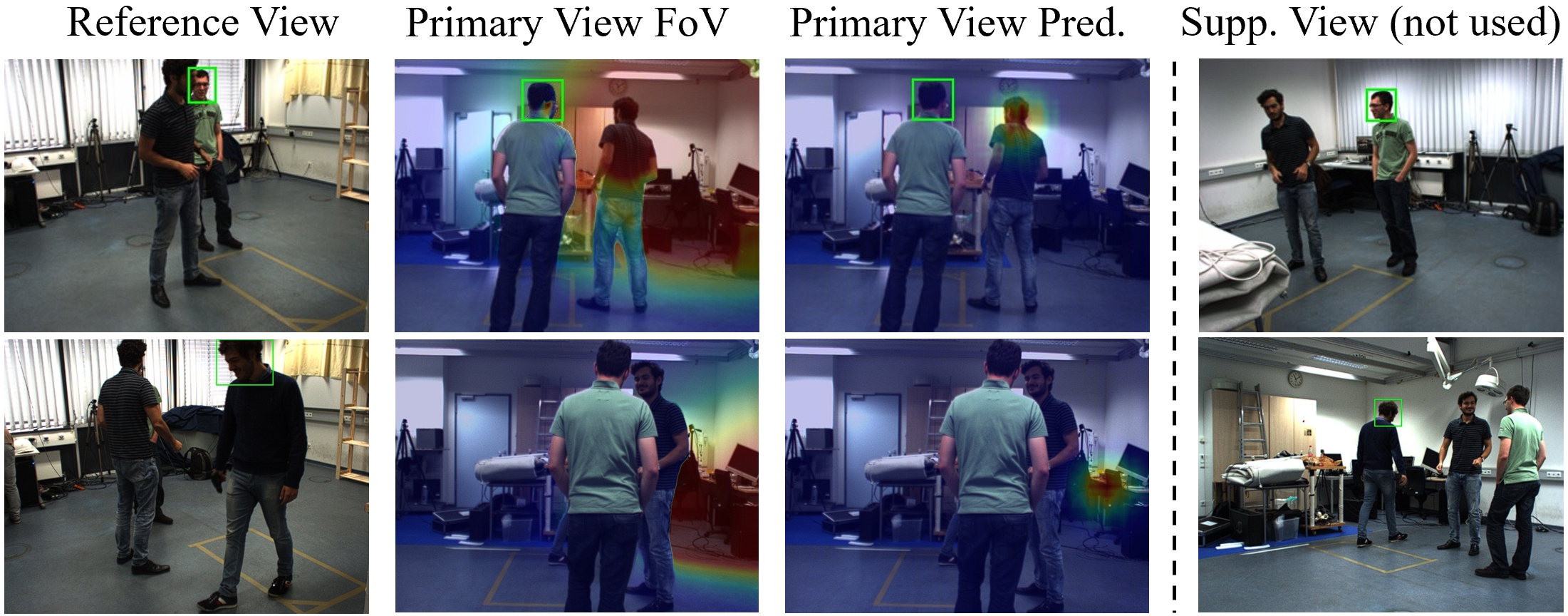}
\caption{Our model evaluated on Shelf dataset. 
Output is shown for the person with an overlayed head box in both general multi-view (Row1) and cross-view (Row2) settings. The rightmost column is just for showing the potential target, and not used as input.} 
\label{fig:run_shelf}
\end{figure}

Regarding the generalization of our method, we demonstrate it in the main paper when evaluated on test scenes not seen in training. In Fig.\ref{fig:run_shelf}, we also apply our models to some example images in the Shelf dataset \cite{pictorial_structure_model}, a multi-view human pose estimation dataset distinct from our dataset. Although no gaze target annotation in available in the dataset, it can be seen from the supplementary view (rightmost column) in the figure that our models predict reasonable locations in both general multi-view and cross-view settings by using the camera parameters provided in the dataset.

While our method is effective in multi-view and cross-view GTE, it still has limitations. The reliance on explicit camera parameters and 3D scene reconstruction in cross-view GTE may limit its applicability in real-world scenarios. Future work could explore learning geometric relationships without camera parameters or performing cross-view GTE without 3D scene reconstruction.

\end{document}